\documentclass{article}

\usepackage{spconf,amsmath,graphicx}
\usepackage{graphicx}
\usepackage{subcaption}
\usepackage{float}
\usepackage{hyperref}
\usepackage[export]{adjustbox}
\usepackage{array}
\usepackage{framed} 
\usepackage{lettrine} 
\hypersetup{
    colorlinks=true,
    linkcolor=blue,
    filecolor=magenta,      
    urlcolor=blue,
}
\usepackage{caption}
\title{Comparative analysis of Automatic Skin Lesion Segmentation with two different implementations}
\name{Md. Kamrul Hasan\textsuperscript{1},  Basel Alyafi\textsuperscript{2},  Fakrul Islam Tushar\textsuperscript{3}}
\address{Joint Master Erasmus program of Medical Imaging and Applications (MAIA)\textsuperscript{1,2,3} \\ University of Girona\textsuperscript{1,2,3}, Spain \\ E-mail: \{kamruleeekuet\textsuperscript{1}, basel931991\textsuperscript{2}, f.i.tushar.eee\textsuperscript{3}\}@gmail.com}

\begin{document}
\maketitle
\begin{abstract}
\lettrine{L}{esion} segmentation from the surrounding skin is the first task for developing automatic Computer-Aided Diagnosis of skin cancer. Variant features of lesion like uneven distribution of color, irregular shape, border and texture make this task challenging. The contribution of this paper is to present and compare two different approaches to skin lesion segmentation. The first approach uses watershed, while the second approach uses mean-shift. Pre-processing steps were performed in both approaches for removing hair and dark borders of microscopic images. The Evaluation of the proposed approaches was performed using Jaccard Index (Intersection over Union or IoU). An additional contribution of this paper is to present pipelines for performing pre-processing and segmentation applying existing segmentation and morphological algorithms which led to promising results.  On average, the first approach showed better performance than the second one with average Jaccard Index over 200 ISIC-2017 challenge images are 89.16 \% and 76.94 \% respectively. 
\end{abstract}
\section{Introduction}
\label{sec:intro}
A skin lesion can be defined as a superficial growth or patch of the skin that does not resemble the area surrounding it. According to the skin cancer foundation, in the past decade (2008-2018), the number of new melanoma cases diagnosed annually increased by 53 percent and one person dies because of melanoma every hour in the world \cite{Kamrulref1}. So, early detection of the skin lesion is one of the foremost challenges in the field of medical engineering. Microscopic (using the microscope) and non-microscopic images, using conventional cameras or smart phones', as shown in Fig \ref{fig:introDermoscopy}, are two types of Medical imaging that are in vivo methods. 
Additionally, skin lesion segmentation is a major step for malignant/benign classification in which doctors tend to recognize some features; lesion asymmetry, border irregularity, area and the timely progress of moles.  
Automated methods challenge due to the wide variety of colors, shapes, sizes, in addition to capturing perturbations, like light reflection, bad illumination, hair and others \cite{ref2}, see Fig \ref{fig:introDermoscopy} (c) and (d).
\begin{framed}
\begin{minipage}[b]{0.35\linewidth}
  \centerline{\includegraphics[width=3.5cm]{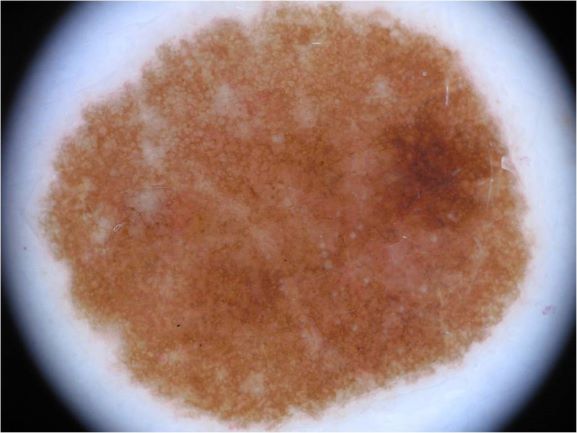}}
  \centerline{(a)}
\end{minipage}
\begin{minipage}[b]{.49\linewidth} 
  \centerline{\includegraphics[width=3.5cm]{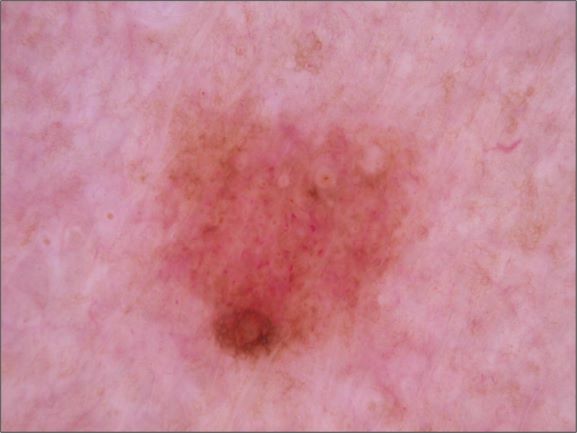}}
  \centerline{(b)}
\end{minipage}
\begin{minipage}[b]{0.48\linewidth}
  \centerline{\includegraphics[width=3.5cm]{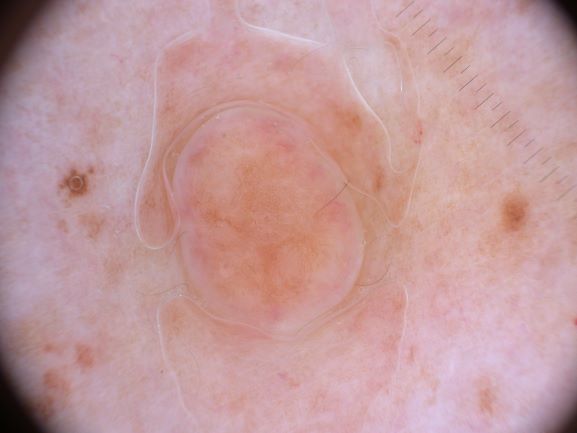}}
  \centerline{(c)} 
\end{minipage}
\begin{minipage}[b]{0.5\linewidth}
  \centerline{\includegraphics[width=3.5cm]{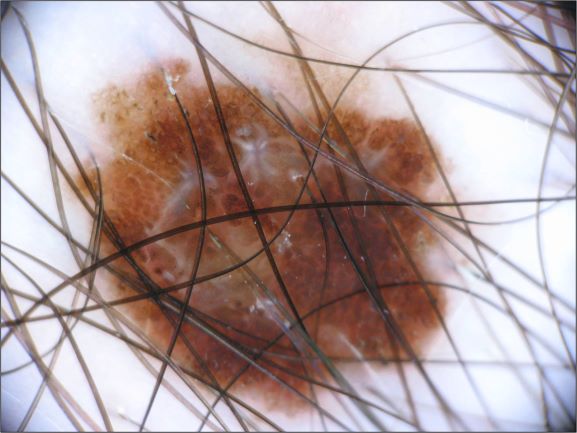}}
  \centerline{(d)}
\end{minipage}
\captionof{figure}{Typical a) dermoscopic (microscopic) image, b) non-microscopic skin lesion image, c) ambiguous lesion, and d) hairy lesion image}
\label{fig:introDermoscopy}
\end{framed}
This work is to segment lesions from 200 images taken from the data base of ISIC 2017 challenge. The segmented lesions are compared with ground truth images and the Jaccard (Intersection over Union or IOU) is calculated.
In \cite{kamrulref2}, color and texture features were extracted from segmented lesion areas. After that, SVM along with k-NN were used for classifications with 61\% F1 score. In \cite{kamrulref3}, thresholding + active contour based method was used with DB1: 10.82\% (XOR); DB2: 13.92\% (XOR). In \cite{kamrulref4}, Hill-climbing algorithm + thresholding were used and that had 94.25\% (TP).
A cutting-edge method has been used by \cite{ref2}, where they proposed a Dense Fully-Convolutional Neural Network (DFCN). By avoiding the redundant computation of neighboring running windows using a dense pooling, they achieved high dice scores in reasonable time.\\ 
This report presents two different methodologies for automatic skin lesion segmentation depending on different color spaces and  is organized as follows:
Section \ref{sec:methodology} briefly shows different steps for all three methods with explanatory flowcharts. Section \ref{sec:results} is dedicated for results and conclusion. 

\section{METhODOLOGY}
\begin{figure}[htb]
    \begin{minipage}[b]{1\linewidth}
    \centerline{\includegraphics[width=7cm]{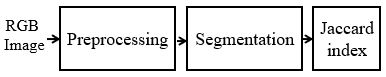}}
    \end{minipage}

    \caption{Overall pipeline}
    \label{fig:k_overall_pipeline}
\end{figure}

As mentioned before, this work included two automatic segmentation methods:
\begin{enumerate}
\item[I.]Method 1 used watershed with RGB color space as the main segmentation algorithm.
\item[II.]Method 2 used meanshift (clustering) and binarization with gray scale images.
\end{enumerate}
The overall pipeline of this paper is shown in Fig. \ref{fig:k_overall_pipeline}.
In the next three subsections, a walk-through explanation for the two methods is provided.
\label{sec:methodology}

\subsection{Method 1: Watershed}
\label{ssec:watershed}

\subsubsection{Pre-processing}
\begin{figure}[htb]    
    \begin{minipage}[b]{1\linewidth}
    \centerline{\includegraphics[width=7.5cm]{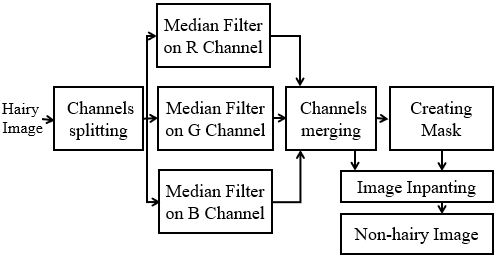}}
    \end{minipage}
    \centerline{(a) Hair removing}\medskip

    \begin{minipage}[b]{1\linewidth}
    \centerline{\includegraphics[width=6cm]{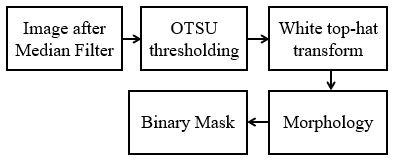}}
    \end{minipage}
    \centerline{(b) Mask creation}\medskip
    
    \begin{minipage}[b]{1\linewidth}
    \centerline{\includegraphics[width=7cm]{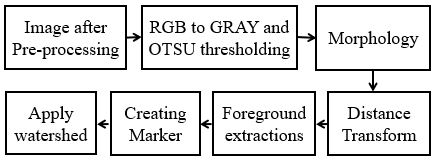}}
    \end{minipage}
  \centerline{(c) Watershed algorithm}\medskip
    \caption{Flow diagram of method 1}
    \label{fig:k_method}
\end{figure}

The pre-processing step is a crucial aspect for the effective analysis of pigmented skin lesions. The main purpose of this step is to improve the quality of lesion region by removing unrelated and surplus parts in the background for further processing. Hair removal from skin lesion images is one of the key problems for the precise segmentation and analysis of the skin lesions. To remove the hair, the block diagram as shown in Fig. \ref{fig:k_method} (a) was used.  
Firstly, median filter was used separately on each color dimension (R,G,B) to remove discontented small particle noise like salt-and-pepper noise. Then, the filtered image was used for creating hair mask to inpaint. The block diagram for creating the mask is shown in Fig. \ref{fig:k_method} (b).
In Fig. \ref{fig:k_method} (b), white top hat has been used with disk (20) type structuring element that can identify hairs. Then binary morphological operations, e.g. closing with a structuring element having size 20 and disk shape, dilation was used to get exact binary mask. To remove the hair, partial differential equation (PDE) and Fast Marching (FM) based inpainting method \cite{kamrulref5}. After carrying out many experiments, 20 neighboring pixels was used for inpainting. Those two methods are well known as NS and TELEA respectively. Another important aspect of this pre-processing pipeline is that it also reduces some other noise, i.e. reflections, shadows, skin lines and air bubbles from the image.
\subsubsection{Segmentation} 
For the segmentation, Watershed was used and its pipeline is shown in Fig. \ref{fig:k_method} (c). To overcome the over segmentation marker-controlled, watershed segmentation has been used \cite{kamrulref7}. After OTSU thresholding, some morphological operation, e.g. opening, and closing, have been done to reduce some white and black noise respectively. To extract the exact sure foreground, distance transform was used that provided a metric or measure of the separation of points in the image. Then marker was created which was used to segment skin lesion using watershed algorithm. Markers based watershed segmentation showed very good results.	
\subsection{Method 2: Mean-shift}
\label{ssec:method2}
\begin{figure}[h]
\centering
\includegraphics[scale=0.65]{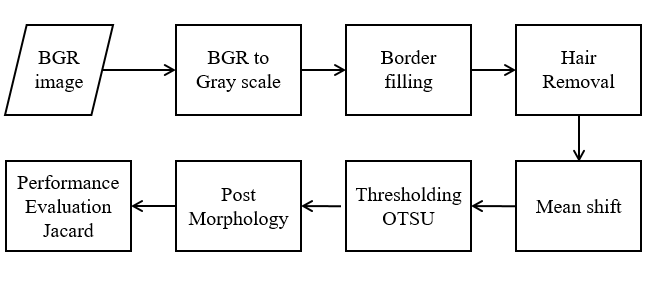}
\caption{Method 2 pipeline}
\label{fig:method2_pipeline}
\end{figure}

Fig \ref{fig:method2_pipeline} provides the bird's-eye view of the methodology used for Method 2.
\subsubsection{ Pre-processing}
\label{sec:method1_pre-processing}
    \begin{itemize}
        \item \textsl{BGR to Gray scale}:
After some experiments, gray scale gave good results in comparison with HSV (H and V channels). Additionally, Gray scale processing provided close results to that given by BGR images. 
    
        \item \textsl{ Border filling}:
The aim of this process is to overcome the effect of the dark borders, in  microscopic images especially, on the thresholding decision.
During this process, binary morphological dilation is applied to remove the one-pixel tiling. Then, a morphology opening is used to block routes leading to the heart of the image (non-border pixels).
After that, region growing is used with four single seeds in the four corners of the image, the goal is to fill pixels with intensities less than a threshold with a pre-defined bright intensity.

        \item \textsl{ Hair removal}:
Hair is one of the artifacts that could be found in a lesion image. To solve this issue, morphology top-hat is used to create the hair mask, then this mask is passed to the inpainting TELEA algorithm along with the gray image to fill hair region defined by the mask with intensities derived from neighborhoods.
end{itemize}
\end{itemize}
\subsubsection{Mean shift and OTSU thresholding}
After Gray-to-BGR transformation is applied (replication is done behind the scene), the result is provided to the pyramids mean shift algorithm which tries to merge pixels close to each other in distance and color spaces. The distance bandwidth and the color bandwidth were tuned to get the required lesion in many cases.
The result of mean-shift is passed to OTSU thresholding, which gives the first unprocessed result which may have some exteriors and gaps in some cases.
\subsubsection{Post morphology}
To ameliorate the problem of exteriors, morphology closing then opening was used to enhance the boundaries of the lesion and remove impairments. 

\section{Results and conclusion}
\label{sec:results}
To evaluate the performance for each method, as mentioned earlier, IoU metric was used. This has been done by counting pixels in intersection region between the result binary image and the provided ground truth and divide it by the number of pixels in the union region.
Overall, results in this paper are divided into three subsections. Subsection \ref{sec:method1_results} is dedicated for Method 1 results, subsection \ref{sec:method2_results} for Method 2, and subsection \ref{sec:comparison_results} is for comparison between method 1 and method 2.  

\subsection{Method 1 results}
\label{sec:method1_results}
The First step, pre-processing, was to reduce noise from the skin image. The outcome is given in Table \ref{tab:method1_noise_removal}.
As can be seen from that table, proposed pipeline not only removes the hair but also reduces ruler noise as well. Another noticeable aspect is if there is no hair nor noise then inpainting keeps this image as it is.   
After removing hair, marker-controlled watershed was used to segment the ROI of the skin lesion. Some of the segmented lesions by the proposed pipeline are given in Table \ref{tab:method1_comparison}.

\begin{table}[h]

\centering
\setlength\tabcolsep{.5pt}
\begin{tabular}{c|c|c}
Original & \multicolumn{1}{|p{2cm}|}{\centering NS \\denoising} & \multicolumn{1}{|p{2cm}|}{\centering TELEA \\ denoising}  \\ \hline

\includegraphics[scale=0.147,valign=c]{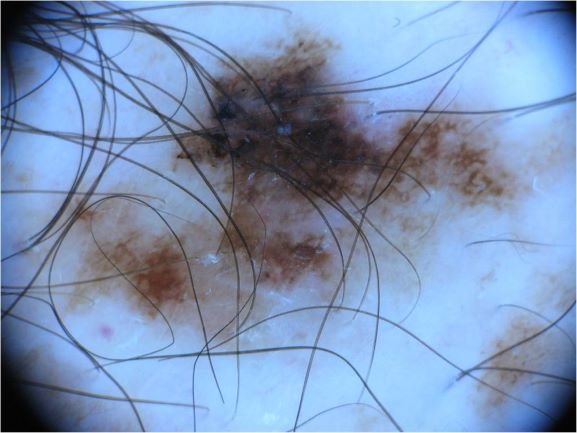}   & \includegraphics[scale=0.063,valign=c]{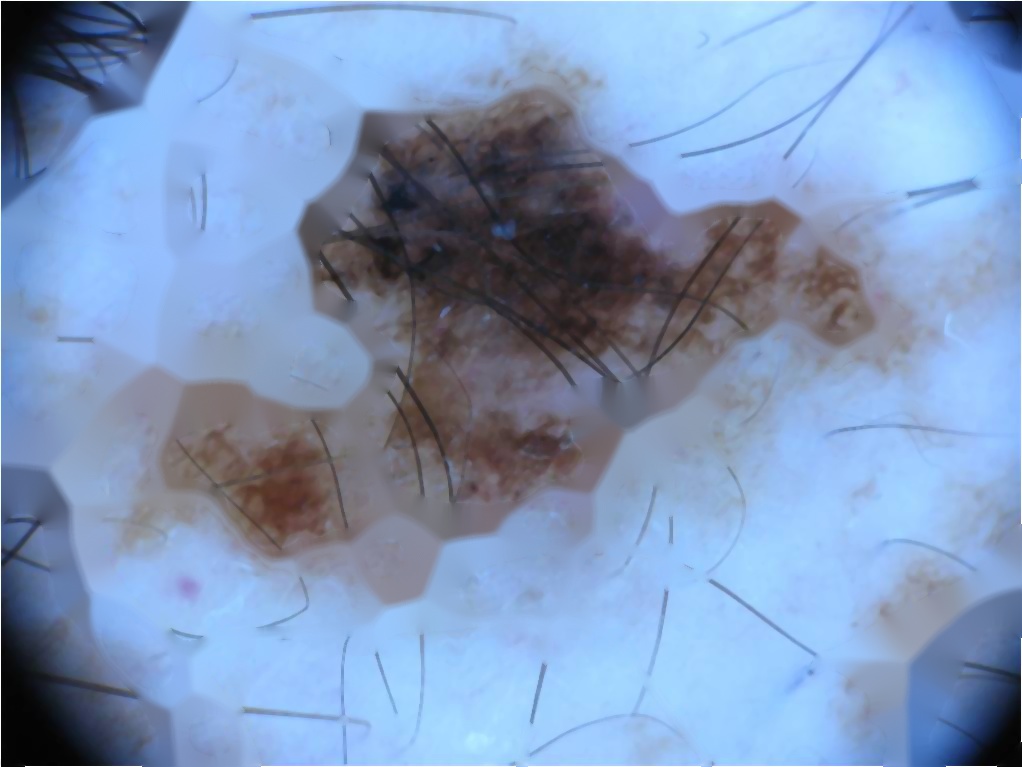} &  
\includegraphics[scale=0.063,valign=c]{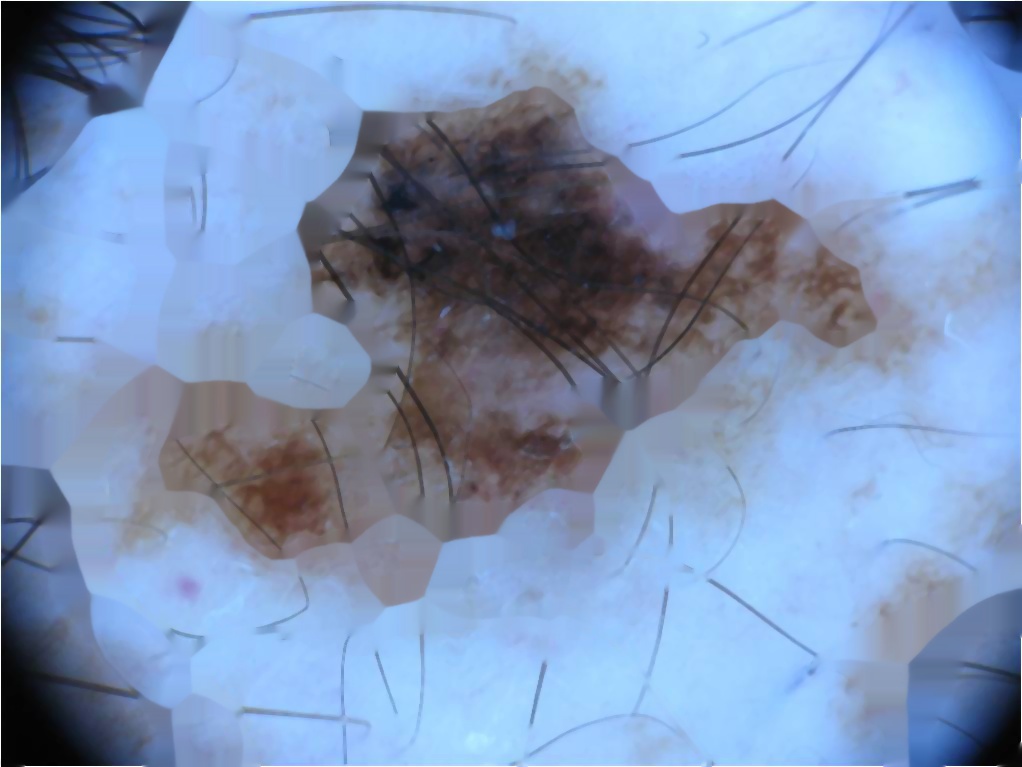}   
\\  \hline

\includegraphics[scale=0.15,valign=c]{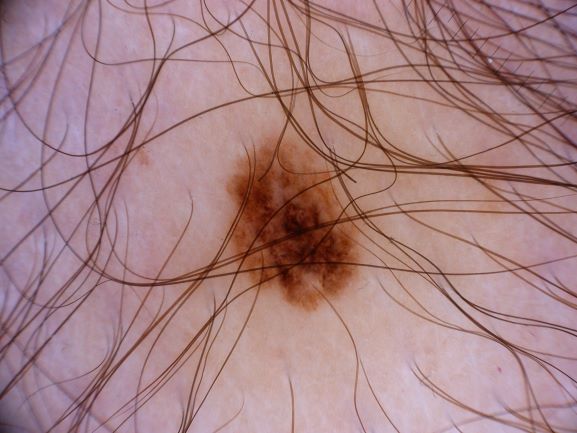}   & \includegraphics[scale=0.031,valign=c]{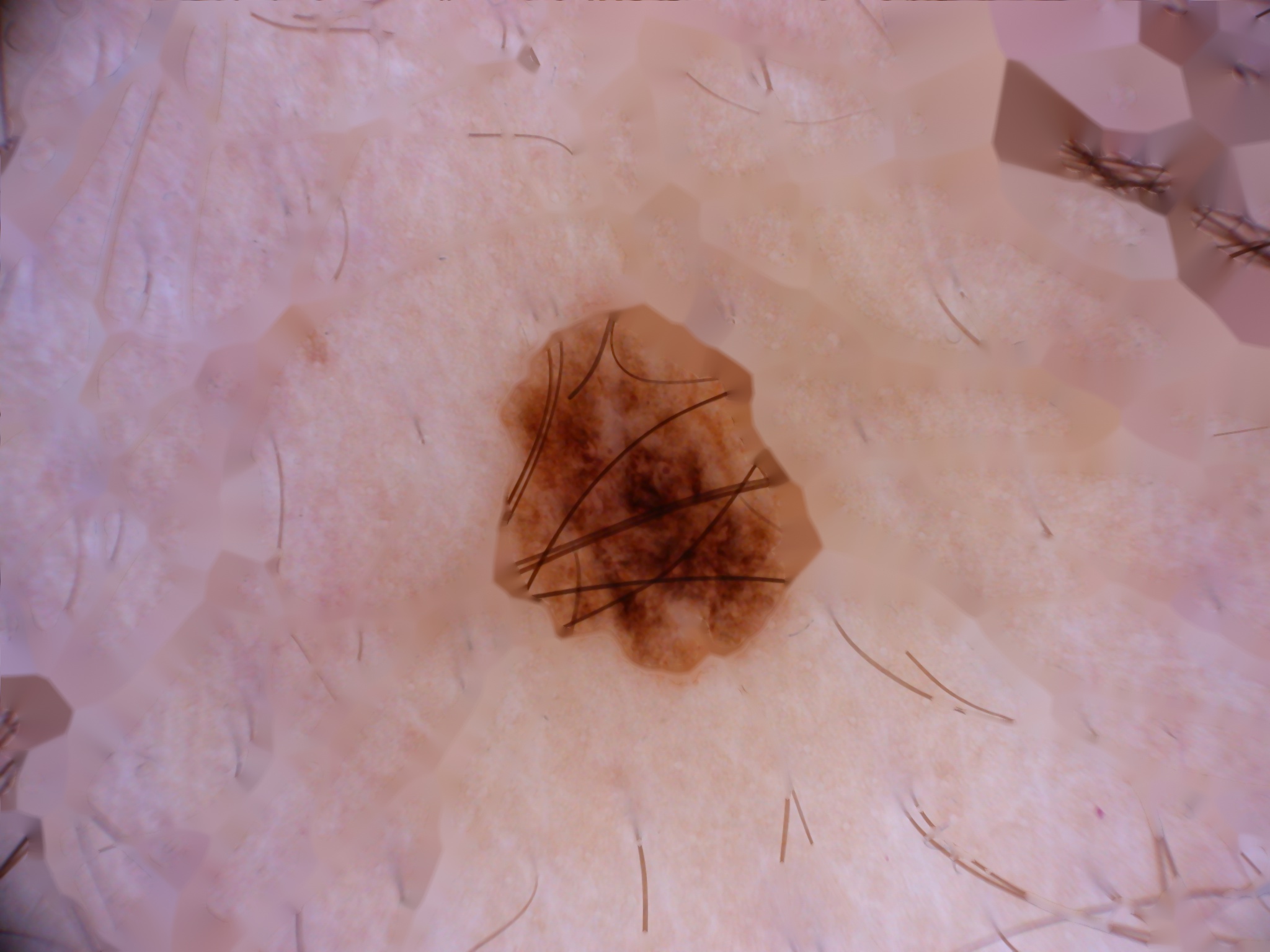} &  
\includegraphics[scale=0.031,valign=c]{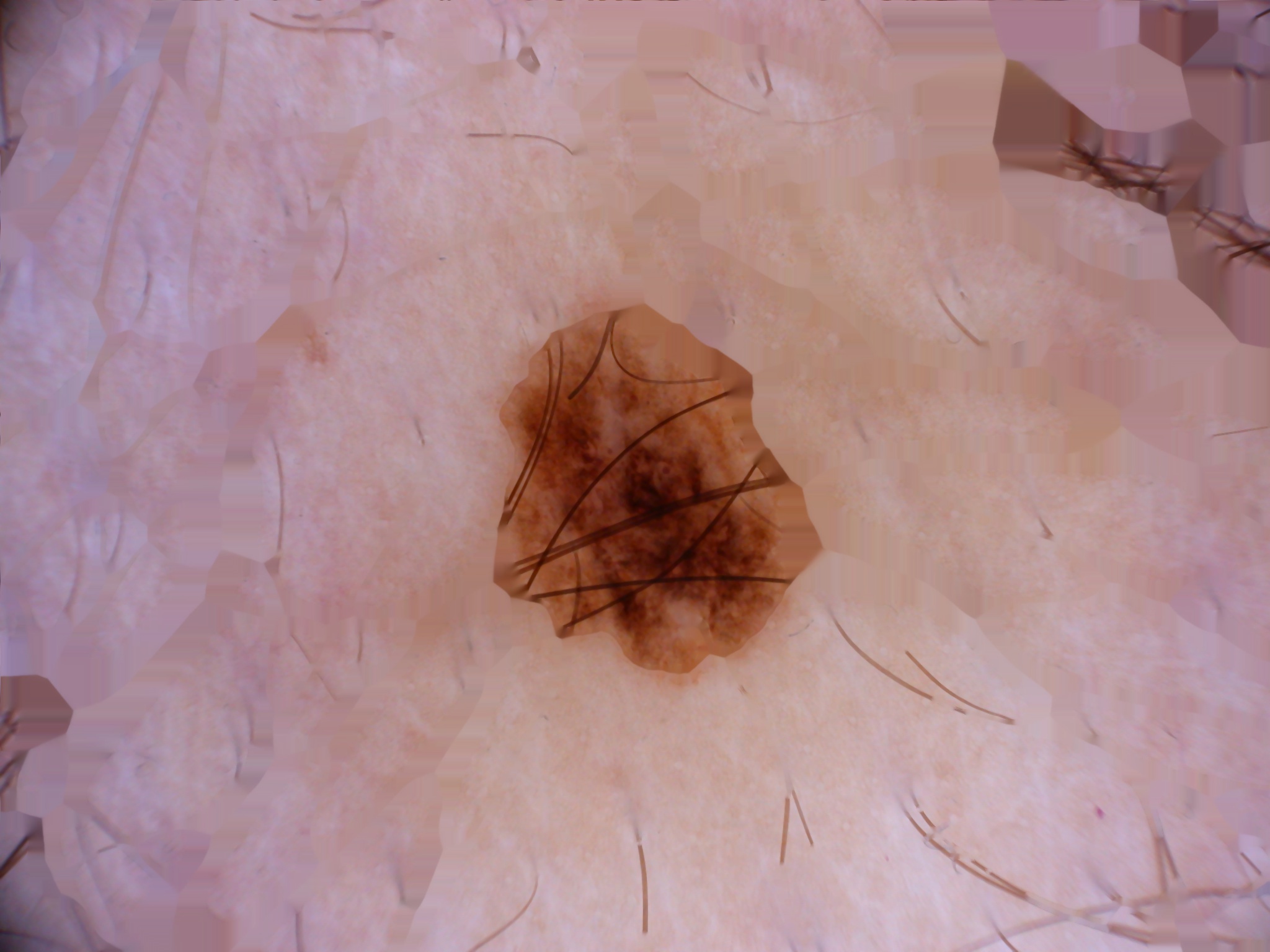}   
\\  \hline

\includegraphics[scale=0.15,valign=c]{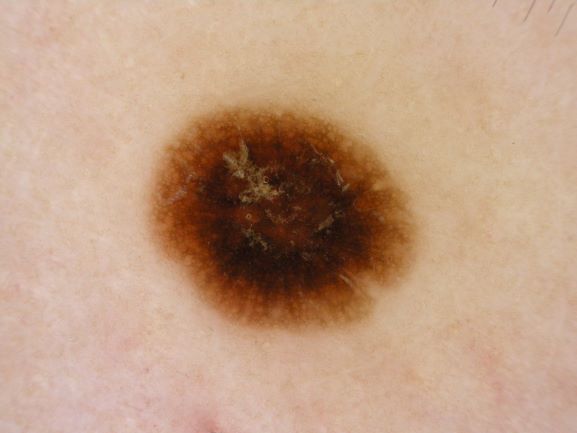}   & \includegraphics[scale=0.025,valign=c]{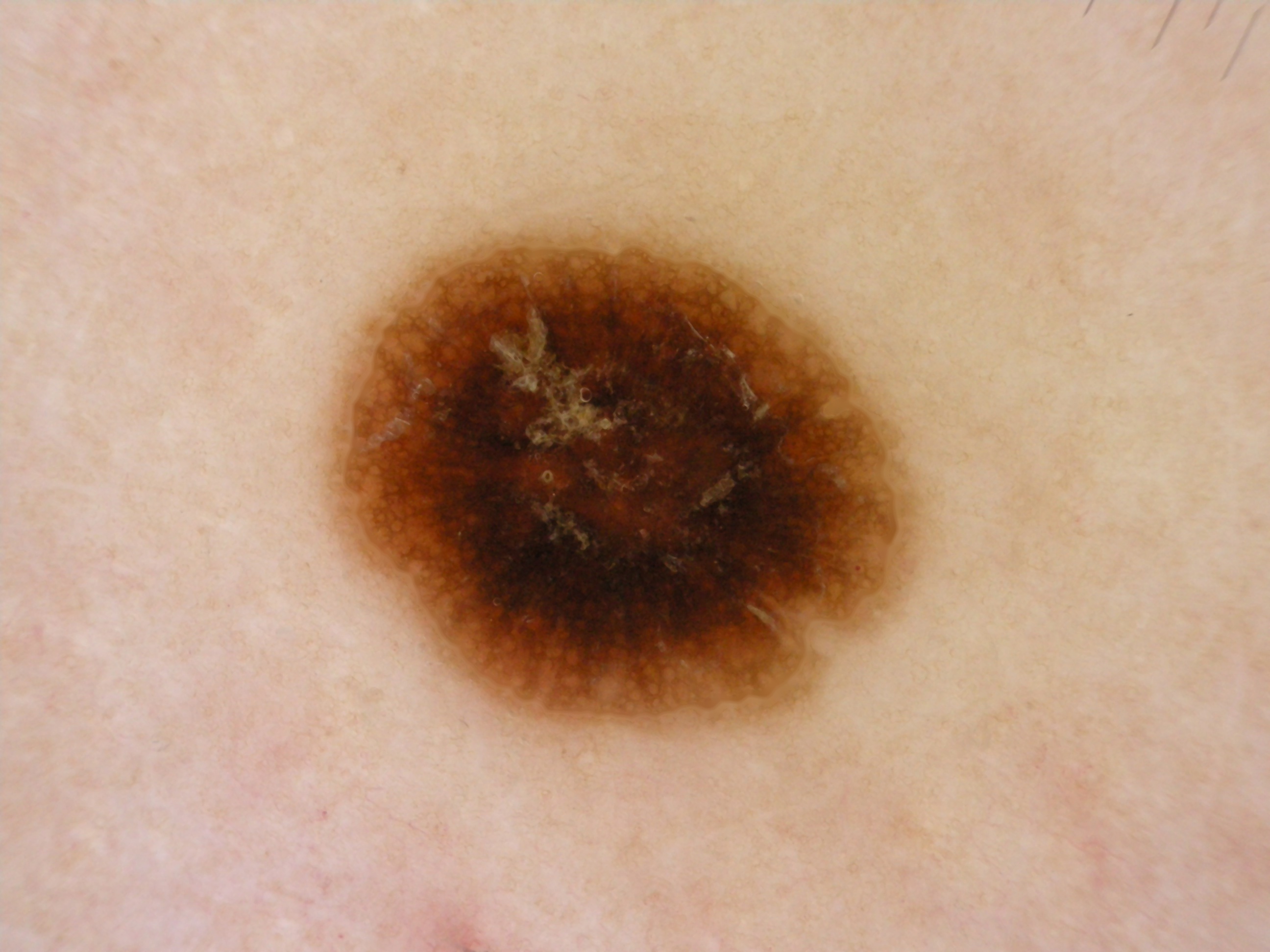} &  
\includegraphics[scale=0.025,valign=c]{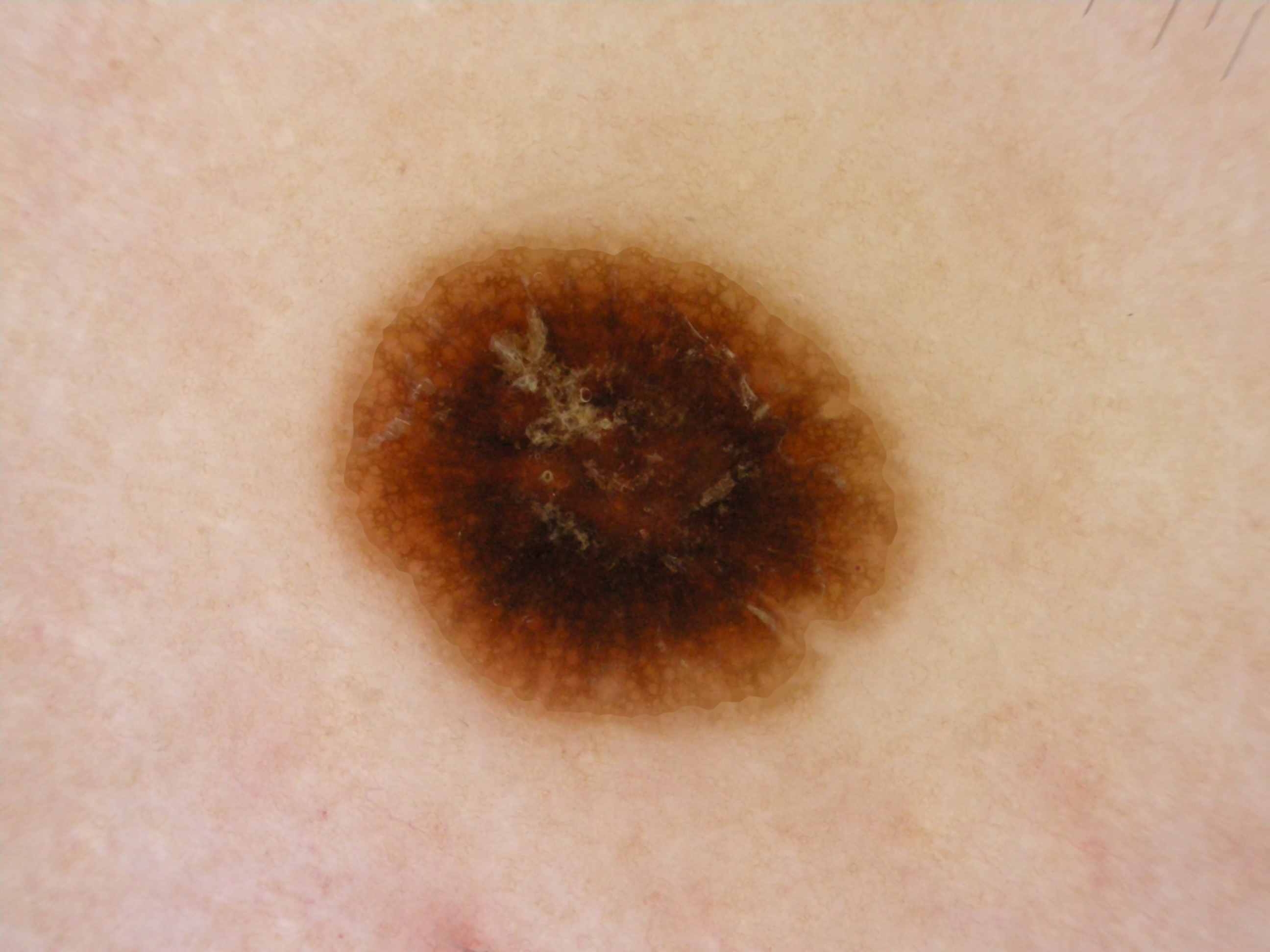}   

\end{tabular}
\caption{Noise removal by two different methods, namely: NS and TELEA.}
\label{tab:method1_noise_removal}
\end{table} 


\begin{table}[h]
\centering

\setlength{\extrarowheight}{1em}
\setlength\tabcolsep{.5pt}
\begin{tabular}{c|c|c|c}
Original &   Groundtruth & \multicolumn{1}{|p{2cm}|}{\centering Seg. after\\ NS} & \multicolumn{1}{|p{2cm}|}{\centering Seg. after \\ TELEA}  \\ \hline

\includegraphics[scale=0.055,valign=c]{images/O-13.jpg}   & \includegraphics[scale=0.055,valign=c]{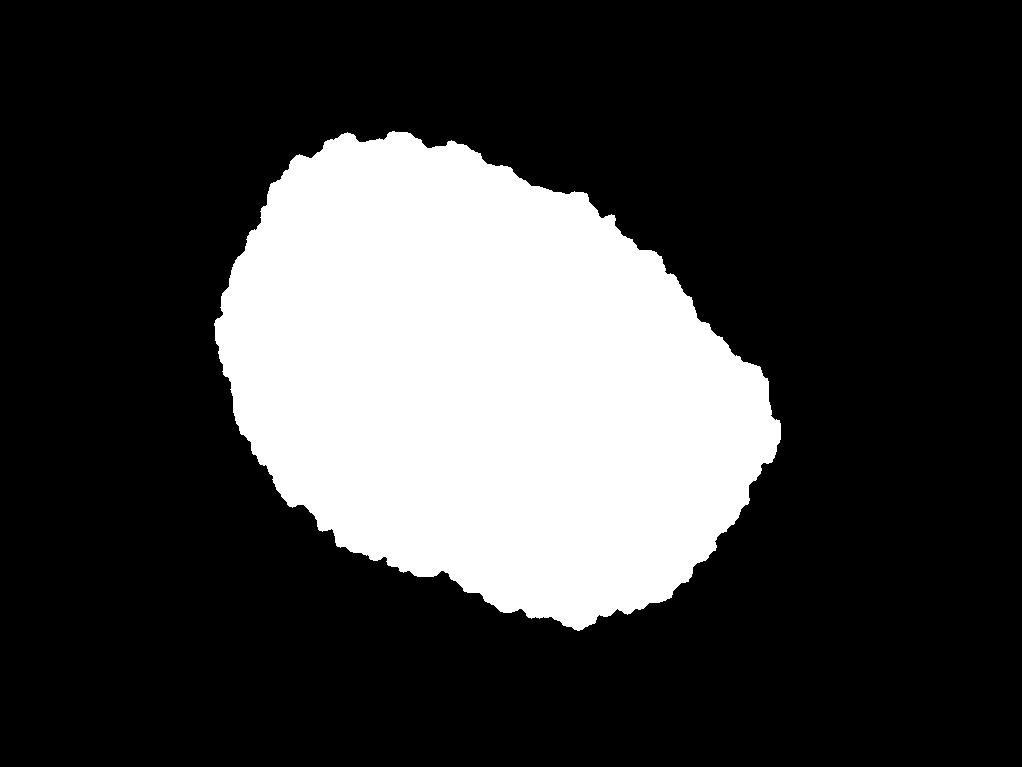} &  
\includegraphics[scale=0.055,valign=c]{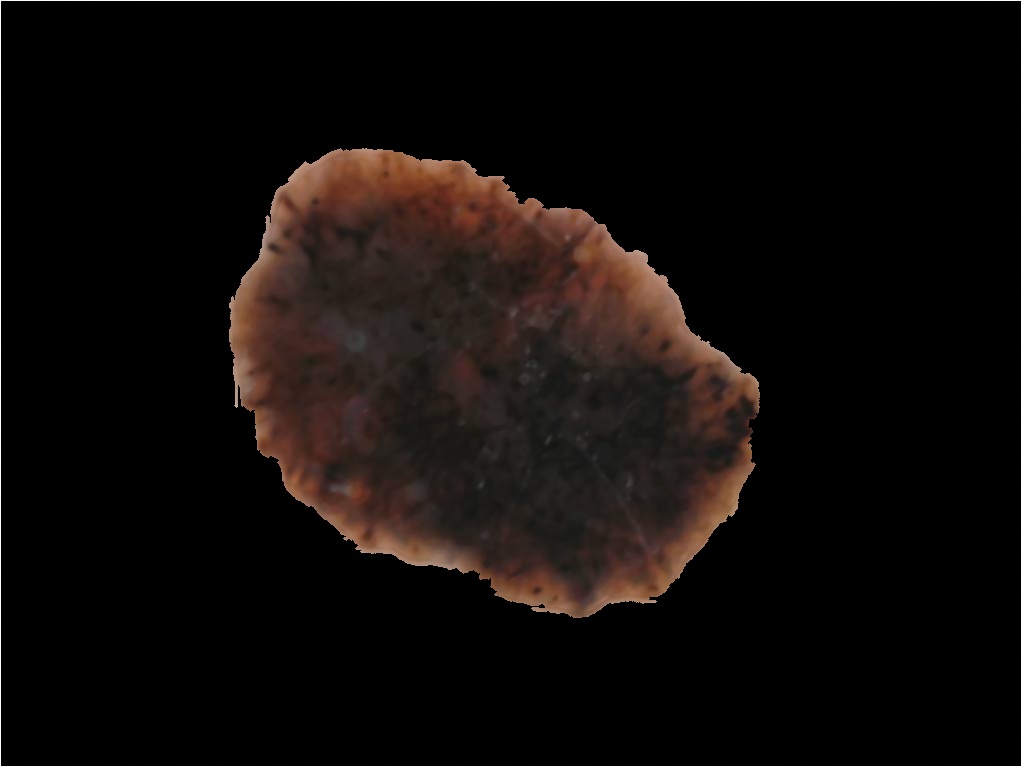}&
\includegraphics[scale=0.055,valign=c]{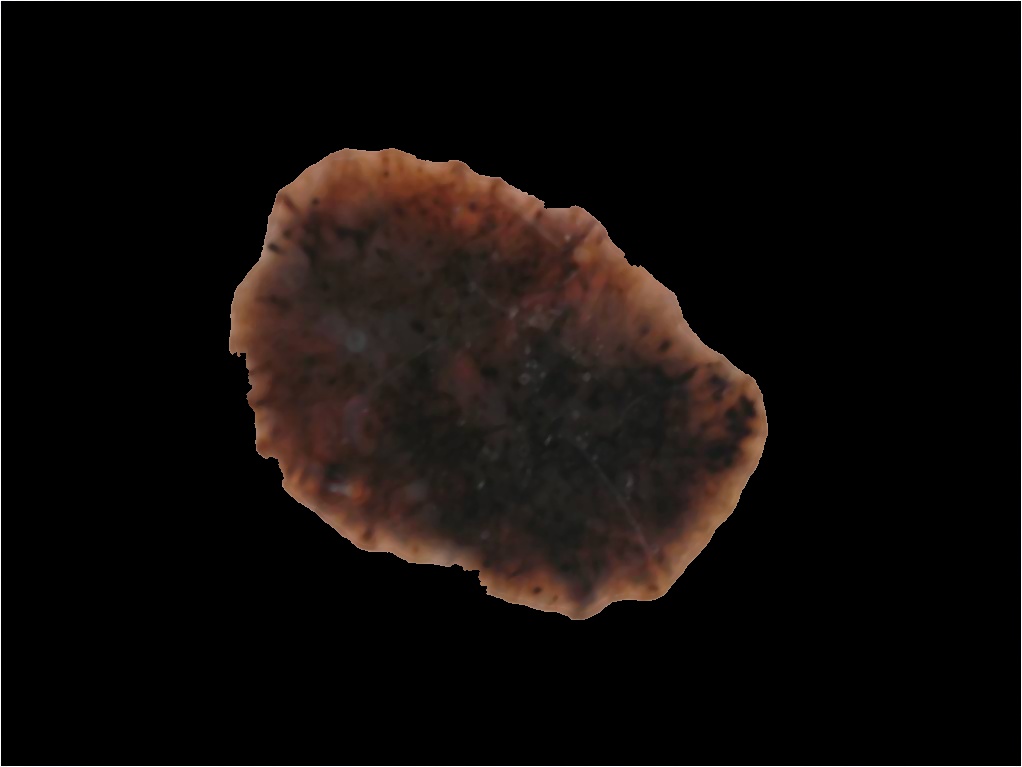}   
\\  \hline

\includegraphics[scale=0.028,valign=c]{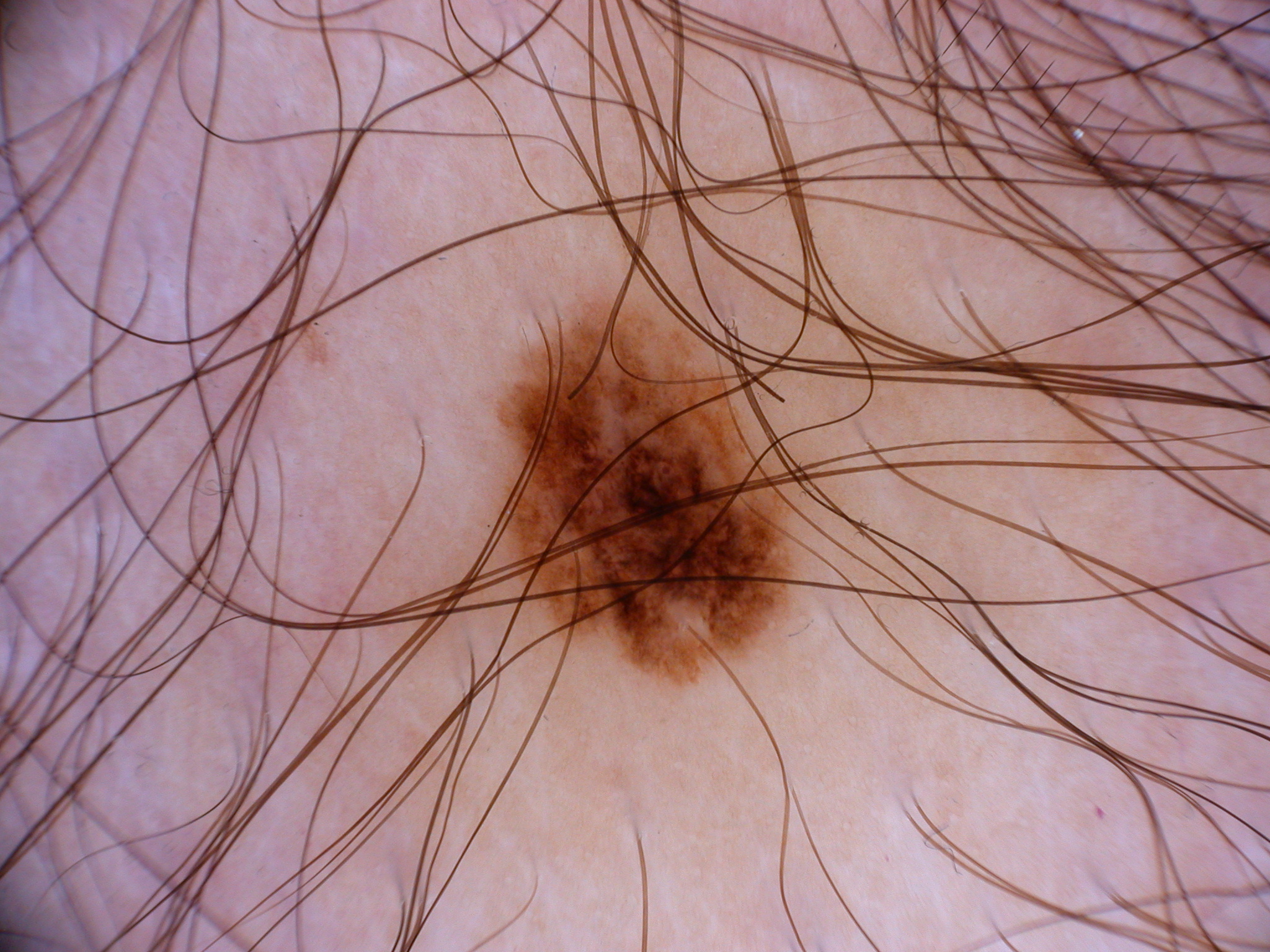}   & \includegraphics[scale=0.028,valign=c]{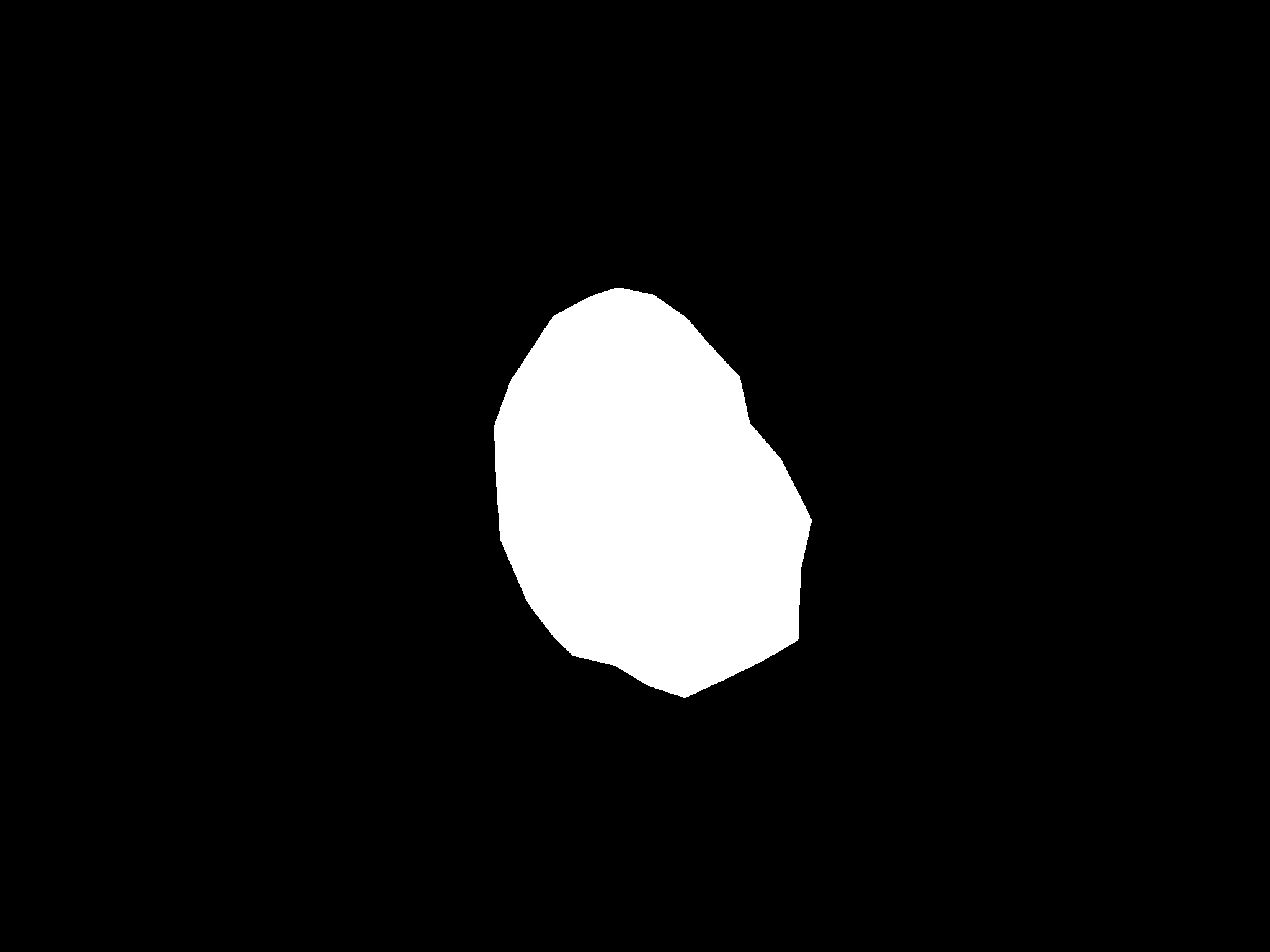} &  
\includegraphics[scale=0.028,valign=c]{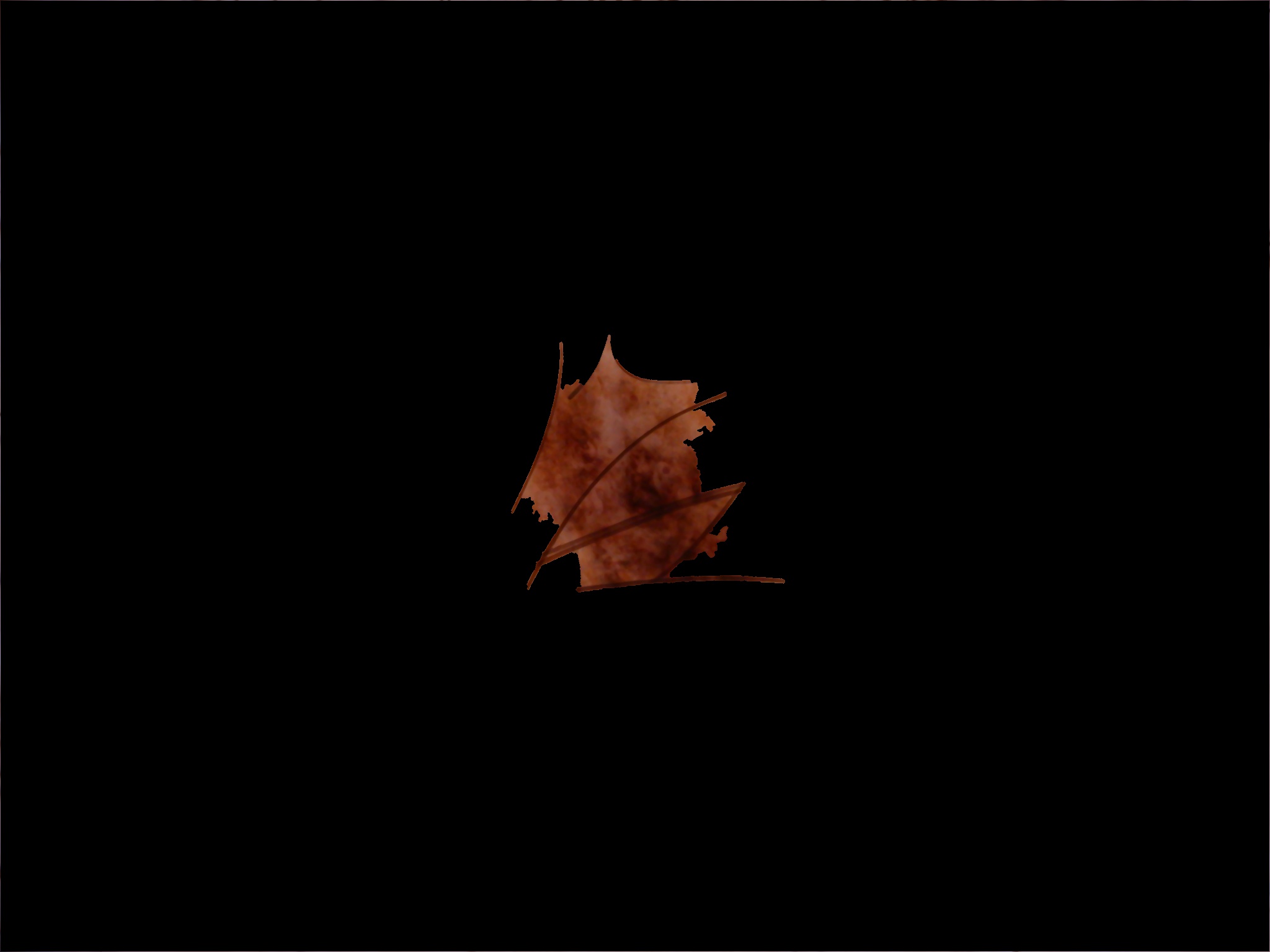}&
\includegraphics[scale=0.028,valign=c]{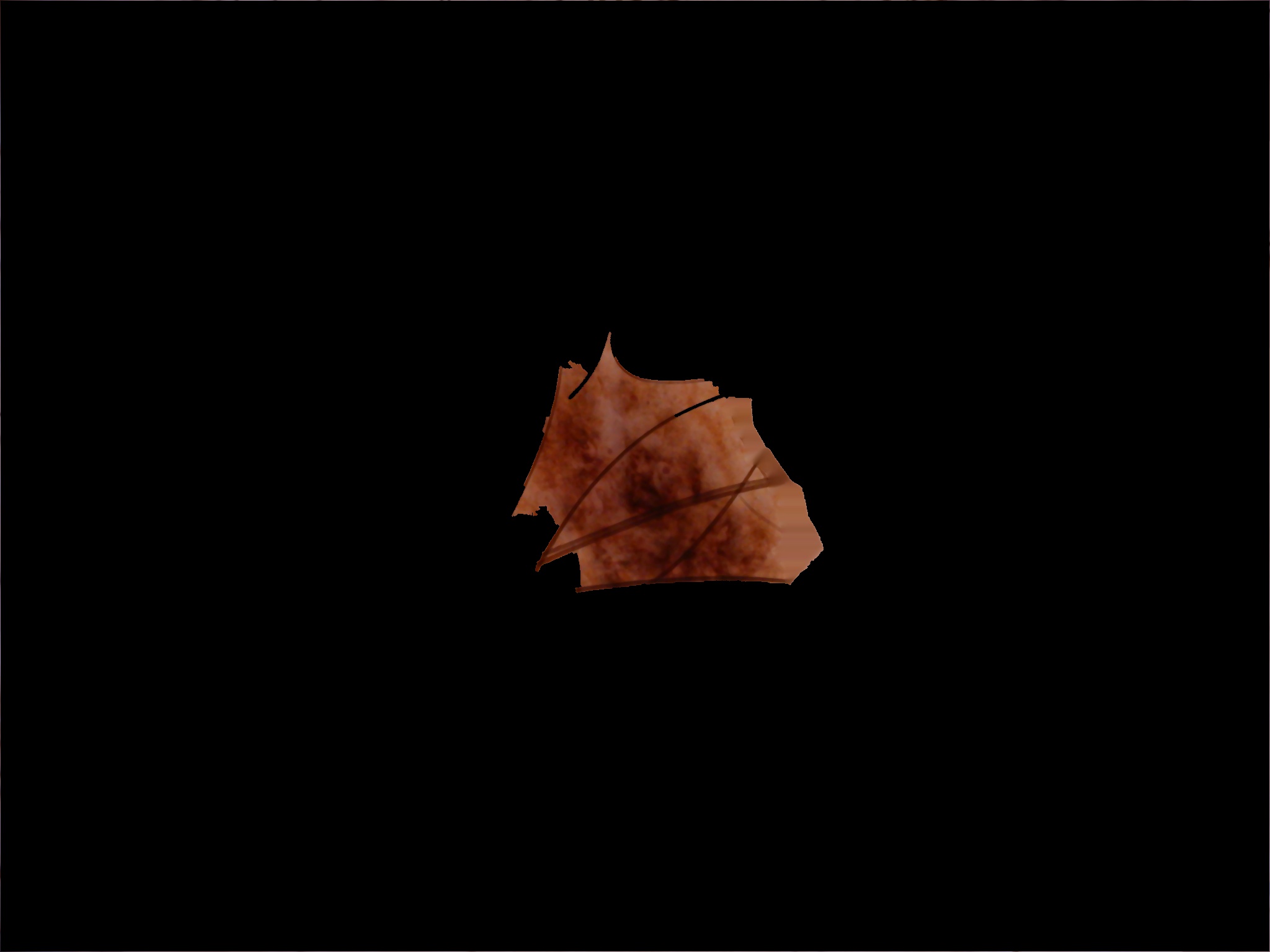}   
\\  \hline

\includegraphics[scale=0.038,valign=c]{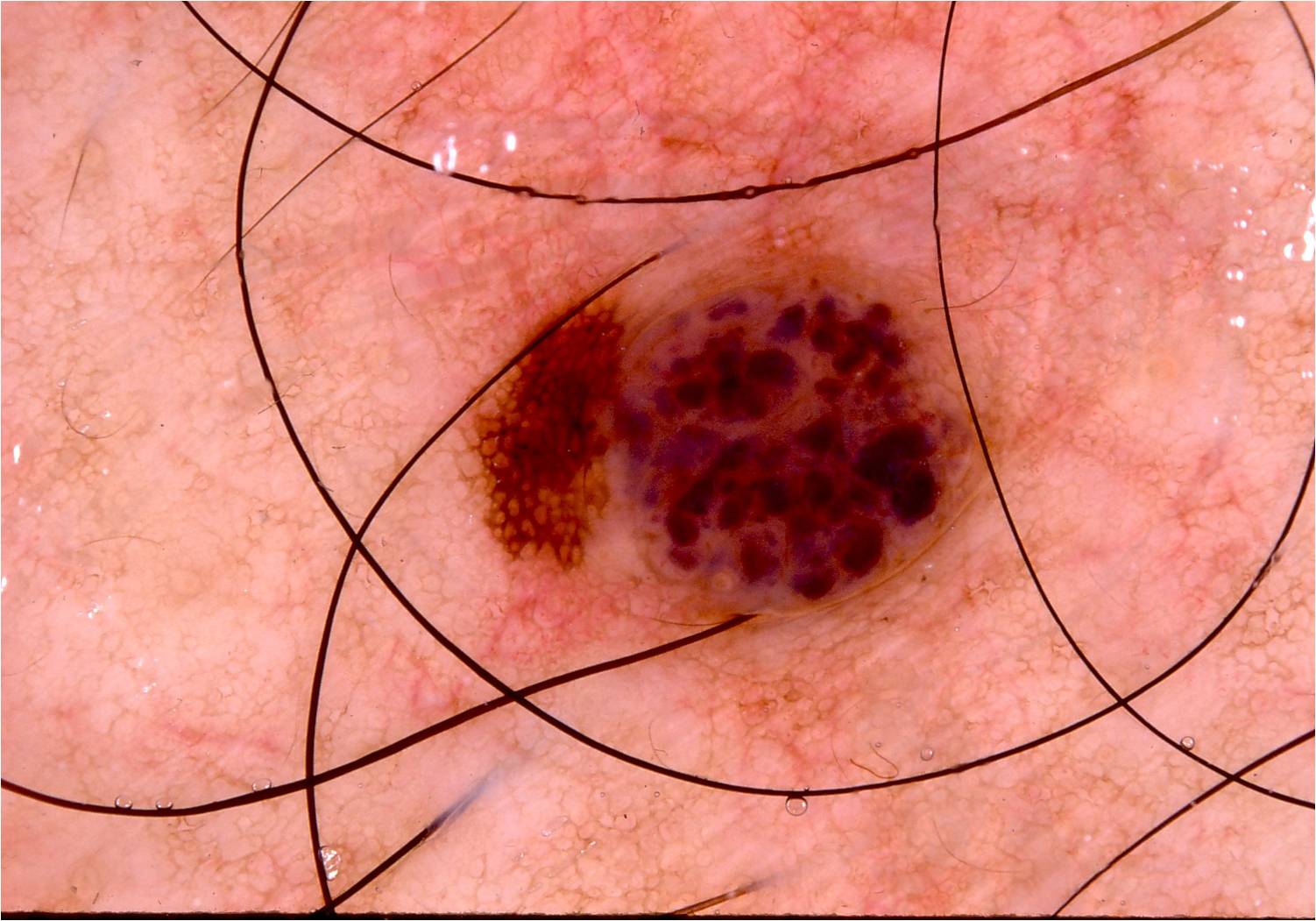}   & \includegraphics[scale=0.038,valign=c]{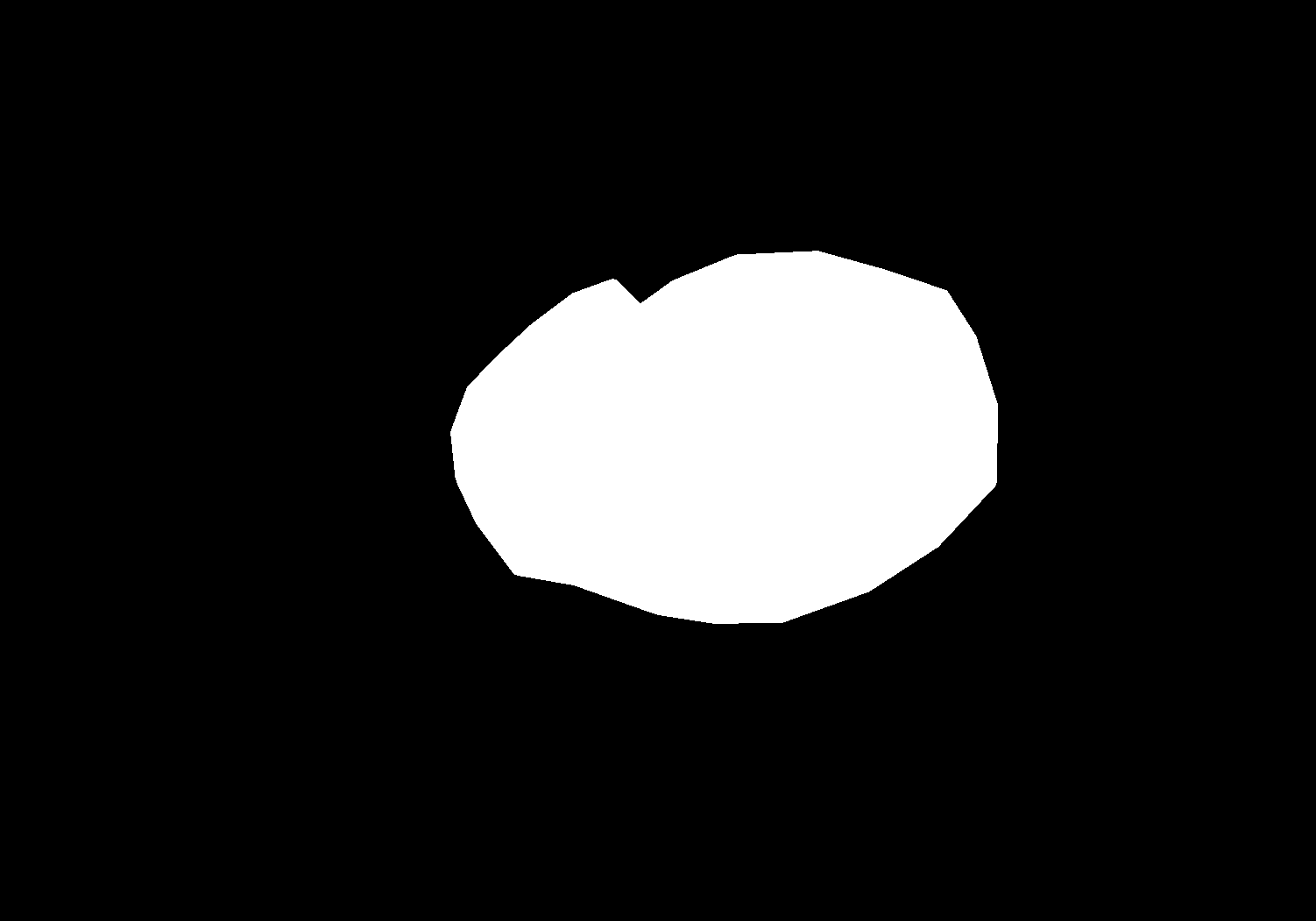} &  
\includegraphics[scale=0.038,valign=c]{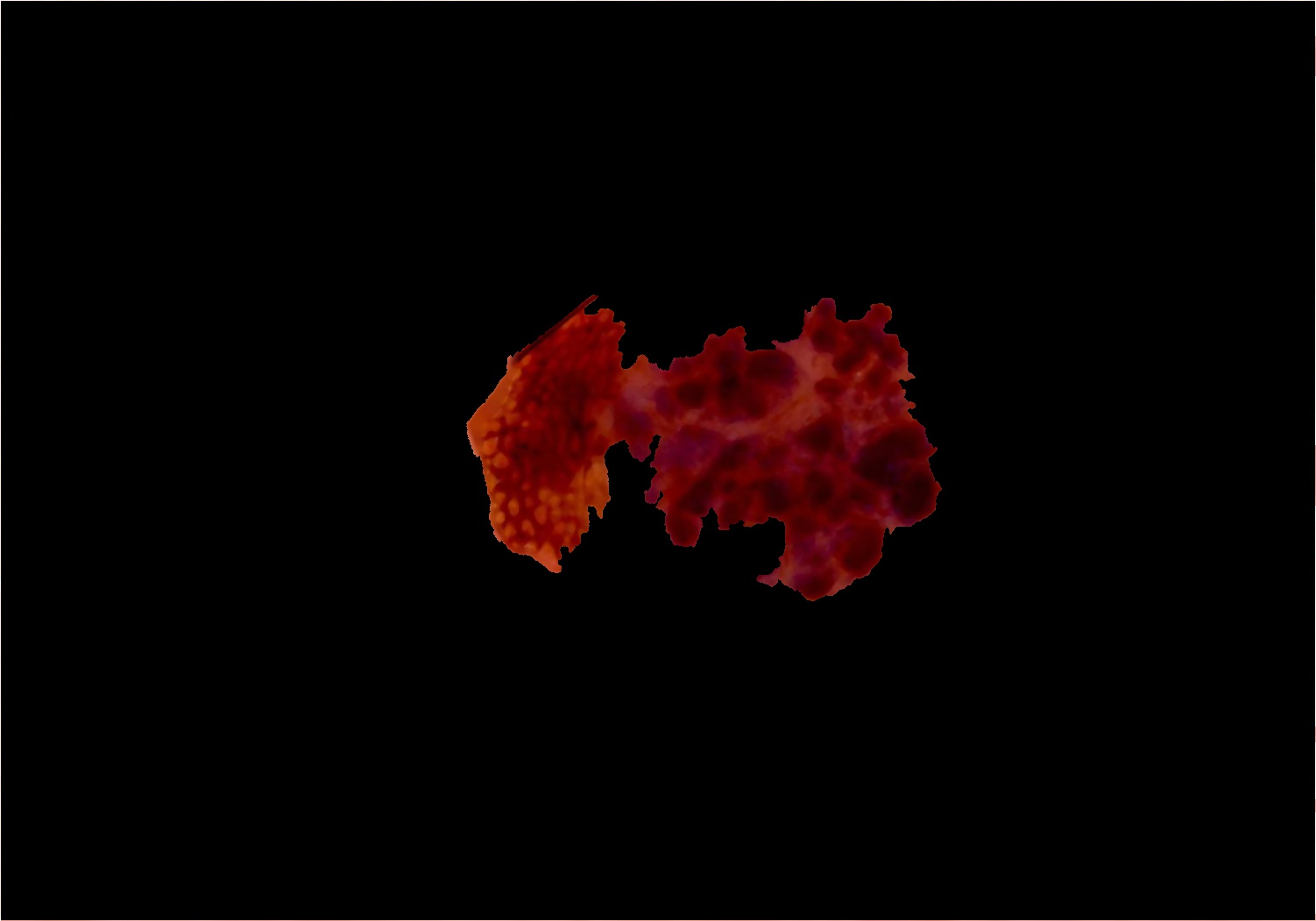}&
\includegraphics[scale=0.038,valign=c]{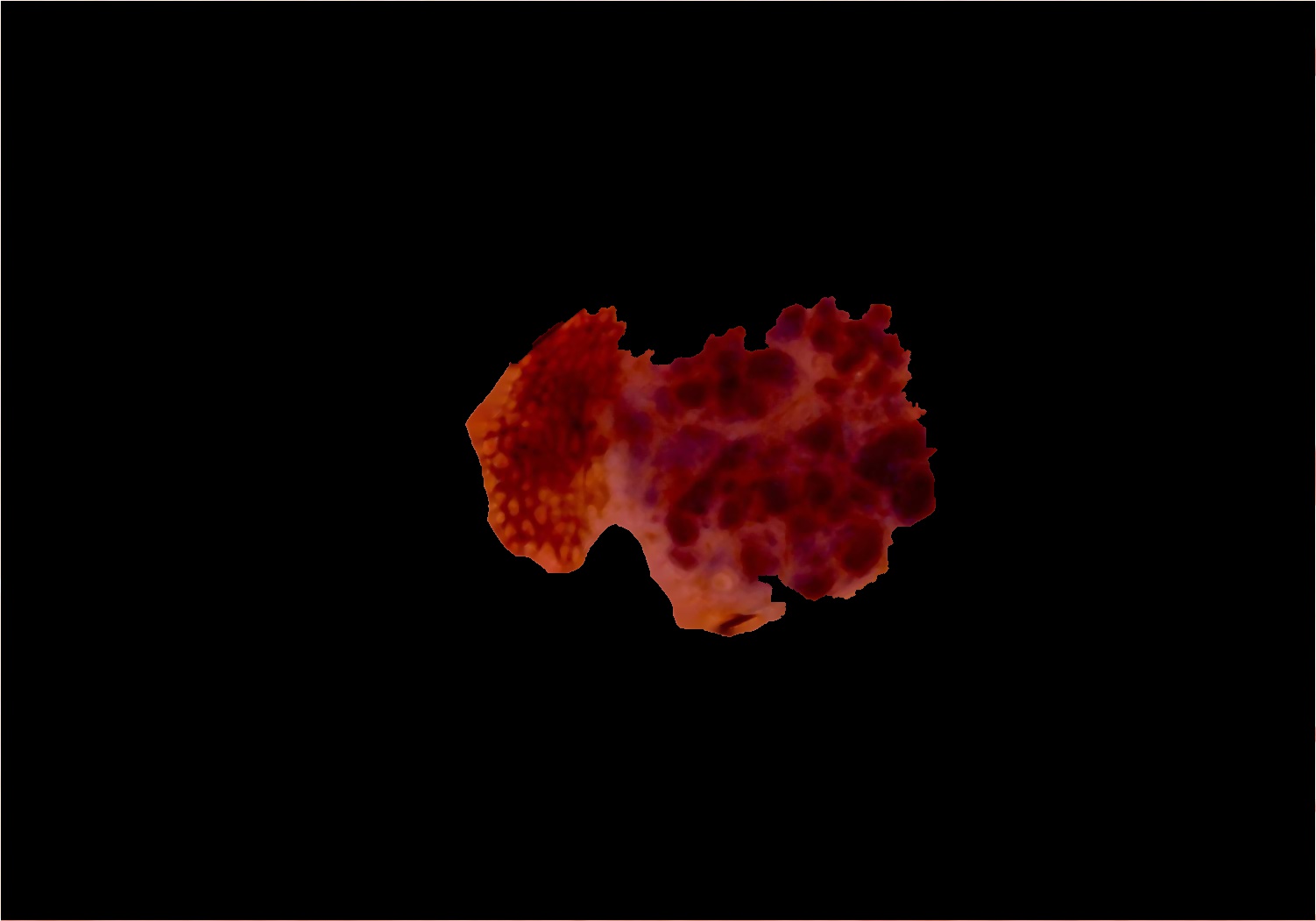}     
\end{tabular}
\caption{Segmentation using watershed after two different pre-processing techniques, namely: NS and TELEA.}
\label{tab:method1_comparison}
\end{table} 
From Table \ref{tab:method1_comparison}, in most of the cases, the segmentation was almost same as ground truth. In some cases especially for the microscopic image performance metric become very less. But over the 200 images, the average Jaccard Index for proposed pipeline are 89.16 \% and 89.08 \% for NS based and TELEA based inpainting in pre-processing respectively.
\subsection{Method 2 results}
\label{sec:method2_results}
This method was applied on BGR images, as well as on gray images. With BGR, an average of 76.94\%, whilst with gray images, it gave a maximum of 74.82\%. This perhaps is a significant outcome, telling that RGB color space carries more information than that can gray images carry. In Table \ref{tab:method2_compare}, it is clearly visible that BGR results are of higher Jaccard, however, it requires more computational resources.

\begin{table}[h]
\setlength\tabcolsep{.5pt}
\begin{tabular}{c|c|c|c|}
Original &  Ground truth & 
\multicolumn{1}{|p{1.8cm}|}{\centering Gray-scale\\ processing} & 
\multicolumn{1}{|p{1.8cm}|}{\centering BGR\\ processing}  \\ \hline

\includegraphics[scale=0.03,valign=c]{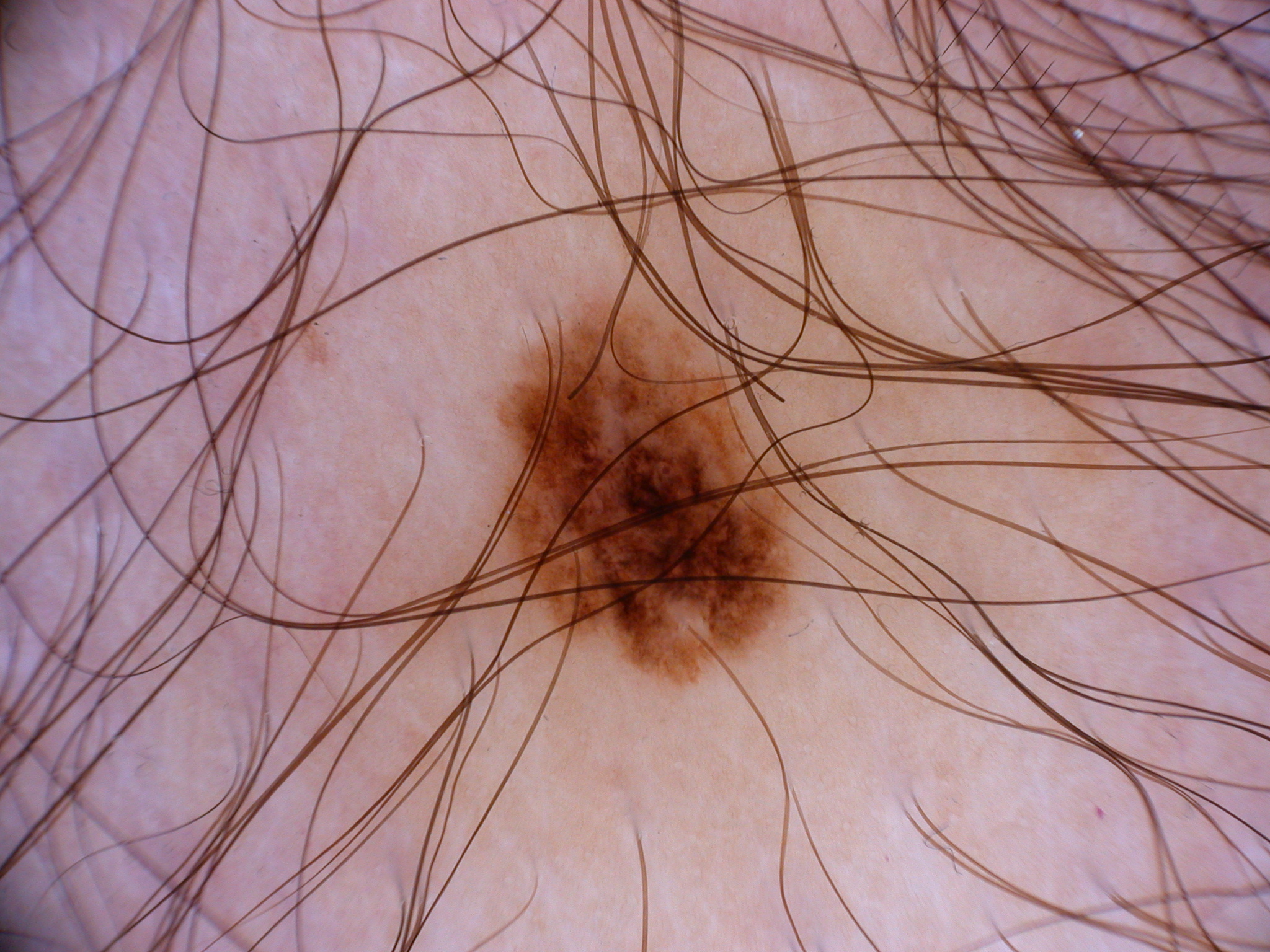}   & 
\includegraphics[scale=0.03,valign=c]{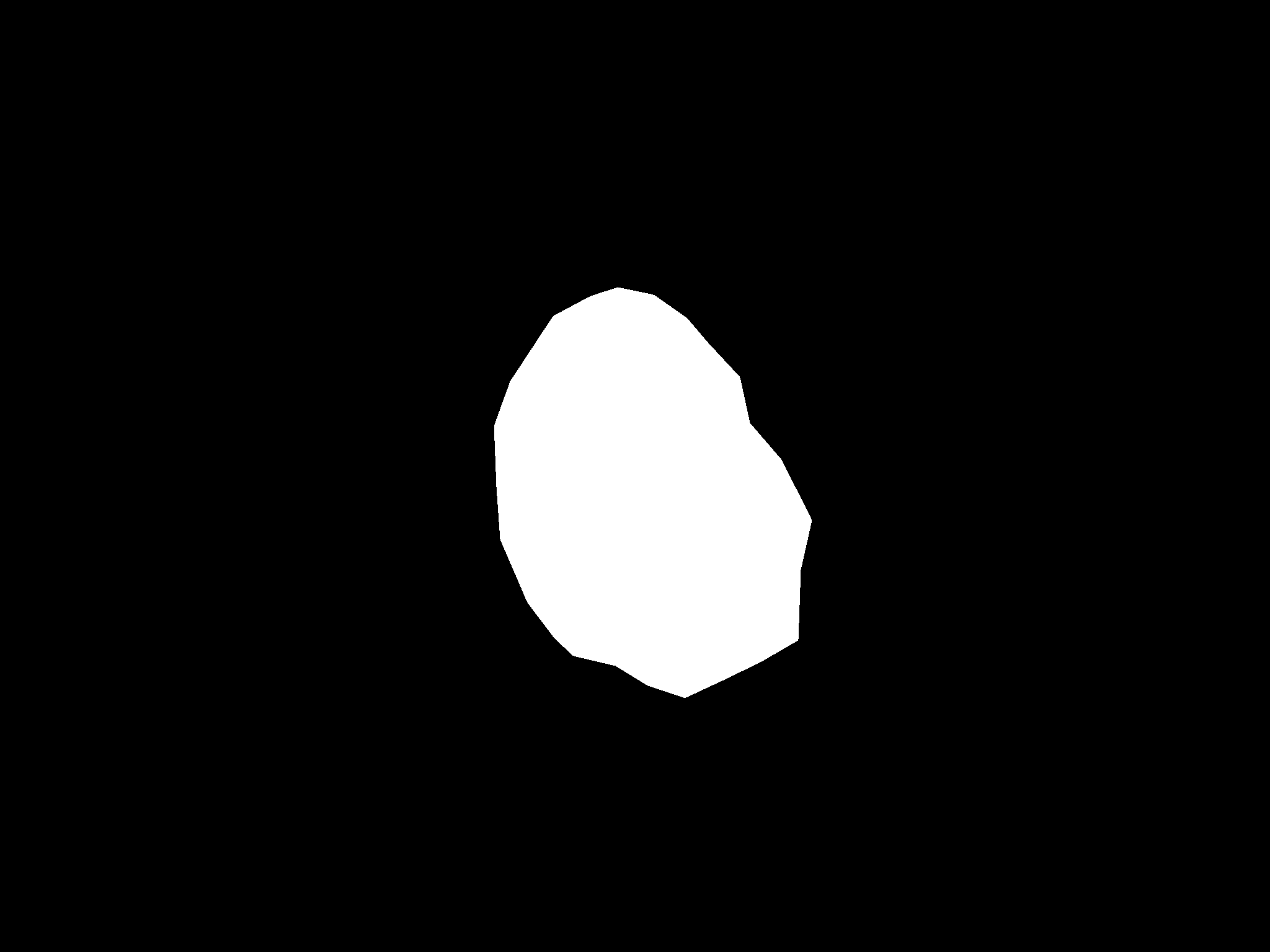}&  
\multicolumn{1}{|p{1.8cm}|}{\centering 
\includegraphics[scale=0.025,valign=c]{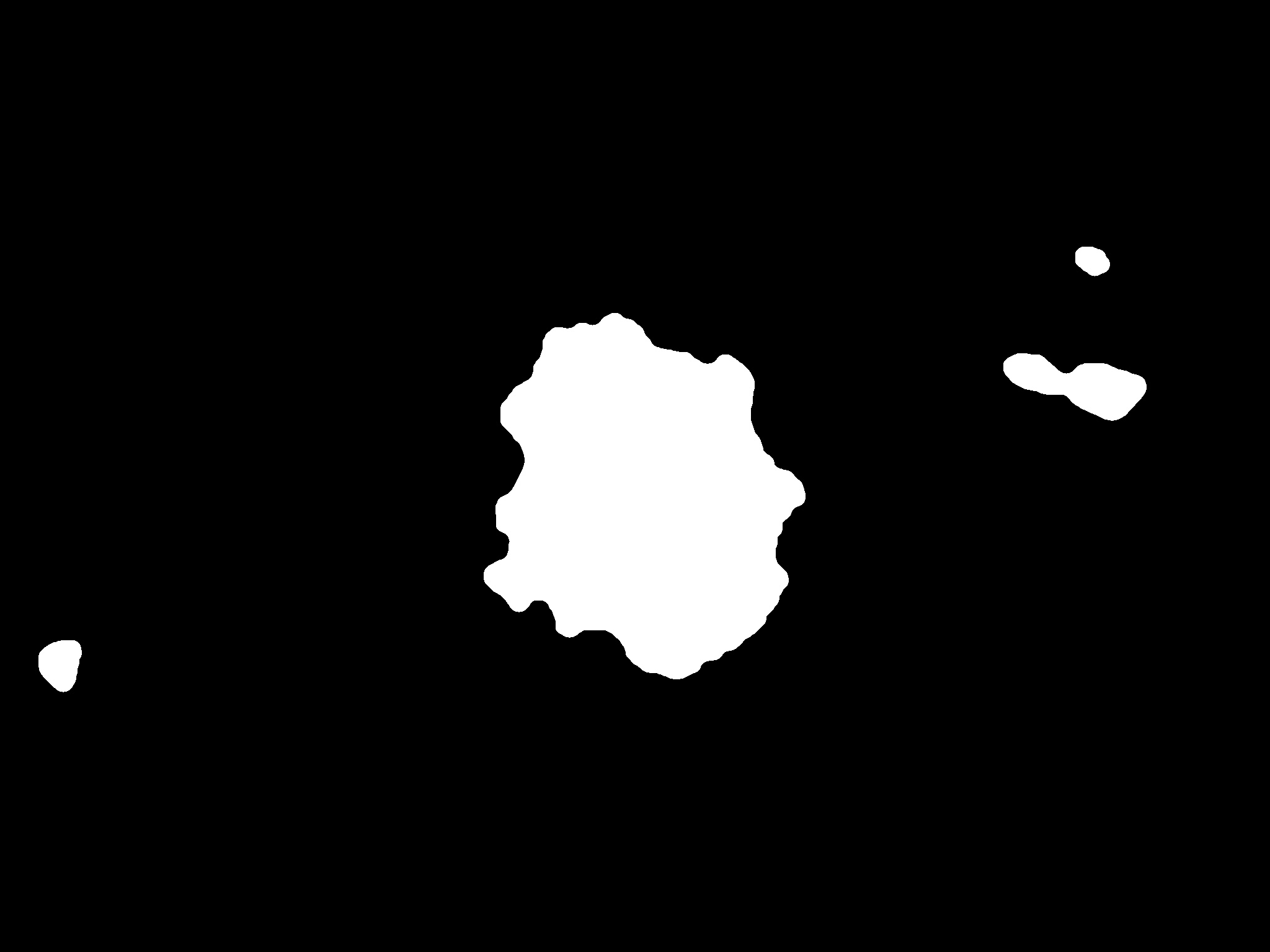}\\ 74.5\%}&
\multicolumn{1}{|p{1.8cm}|}{\centering 
\includegraphics[scale=0.025,valign=c]{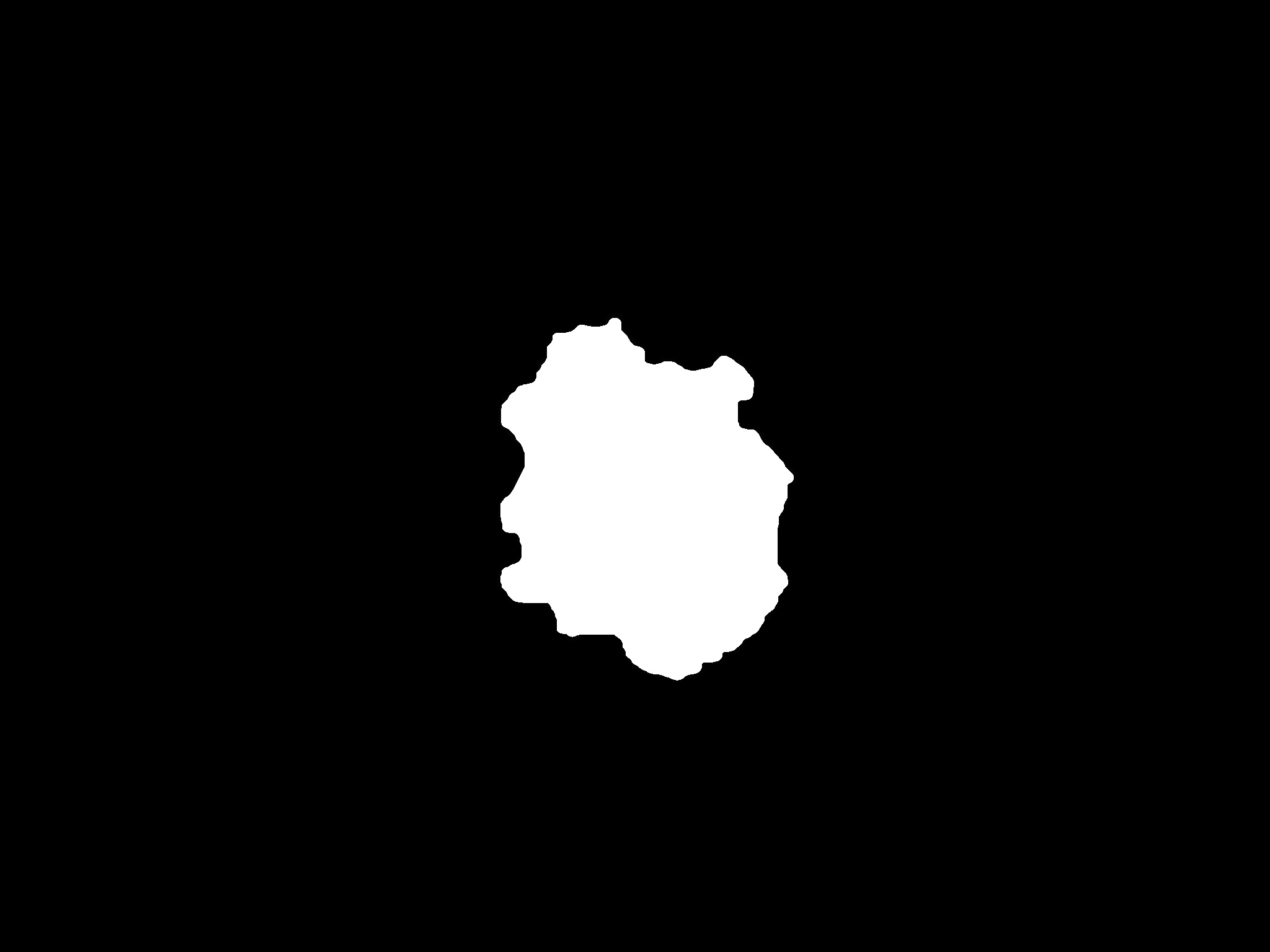}\\ 80.22\%} \\ \hline  

\includegraphics[scale=0.022,valign=c]{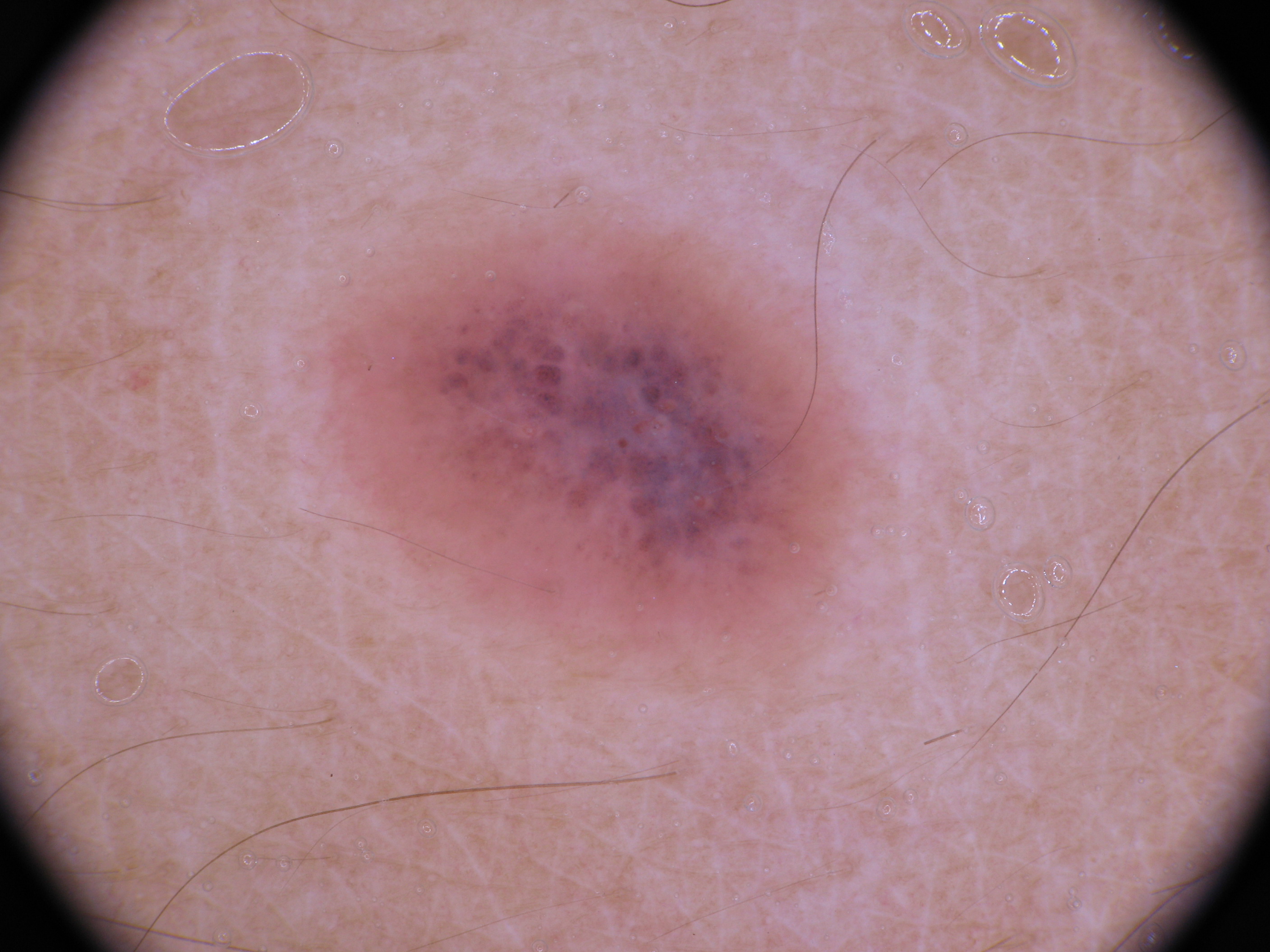}   & 
\includegraphics[scale=0.022,valign=c]{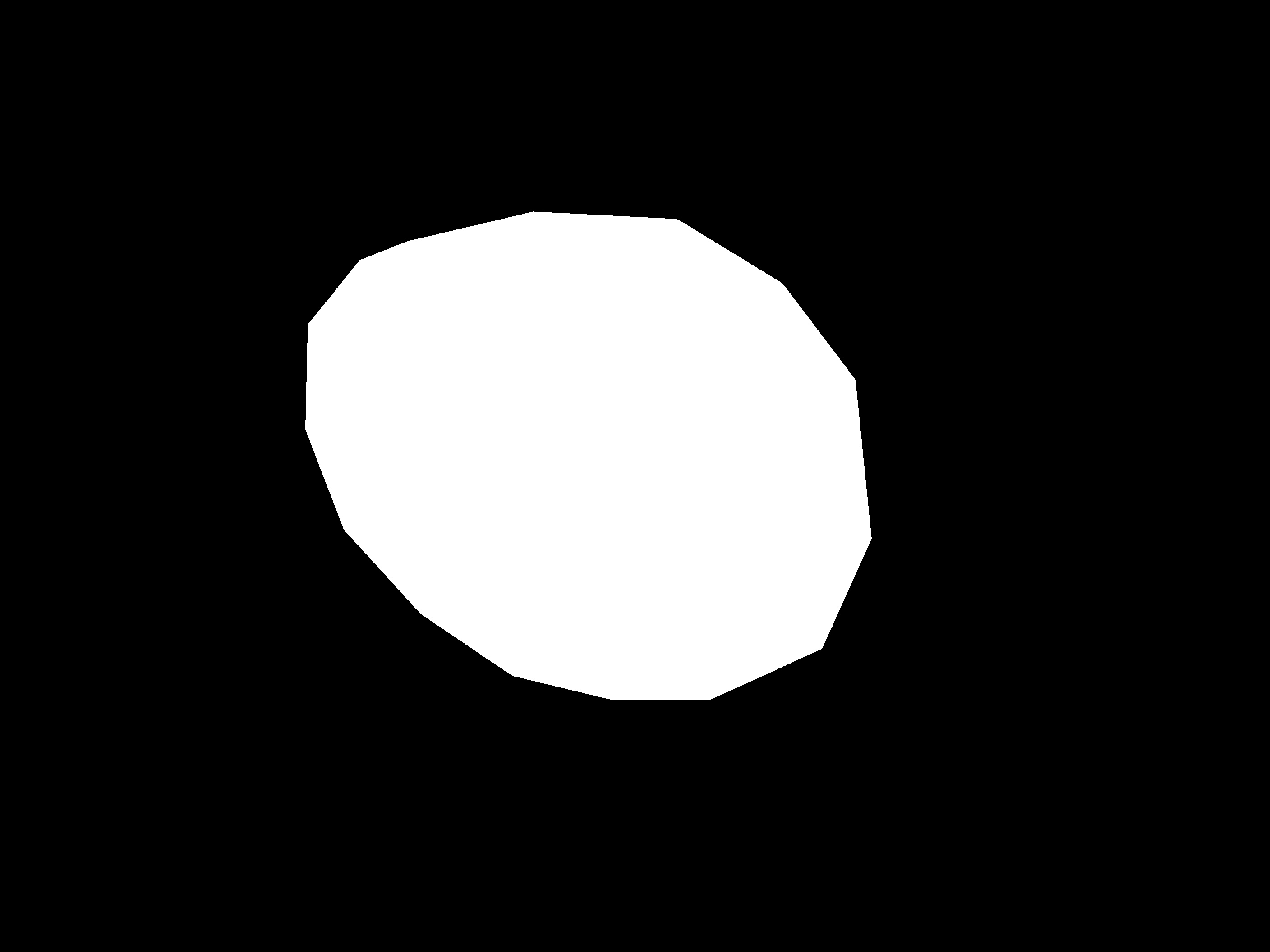}&  
\multicolumn{1}{|p{1.8cm}|}{\centering 
\includegraphics[scale=0.018,valign=c]{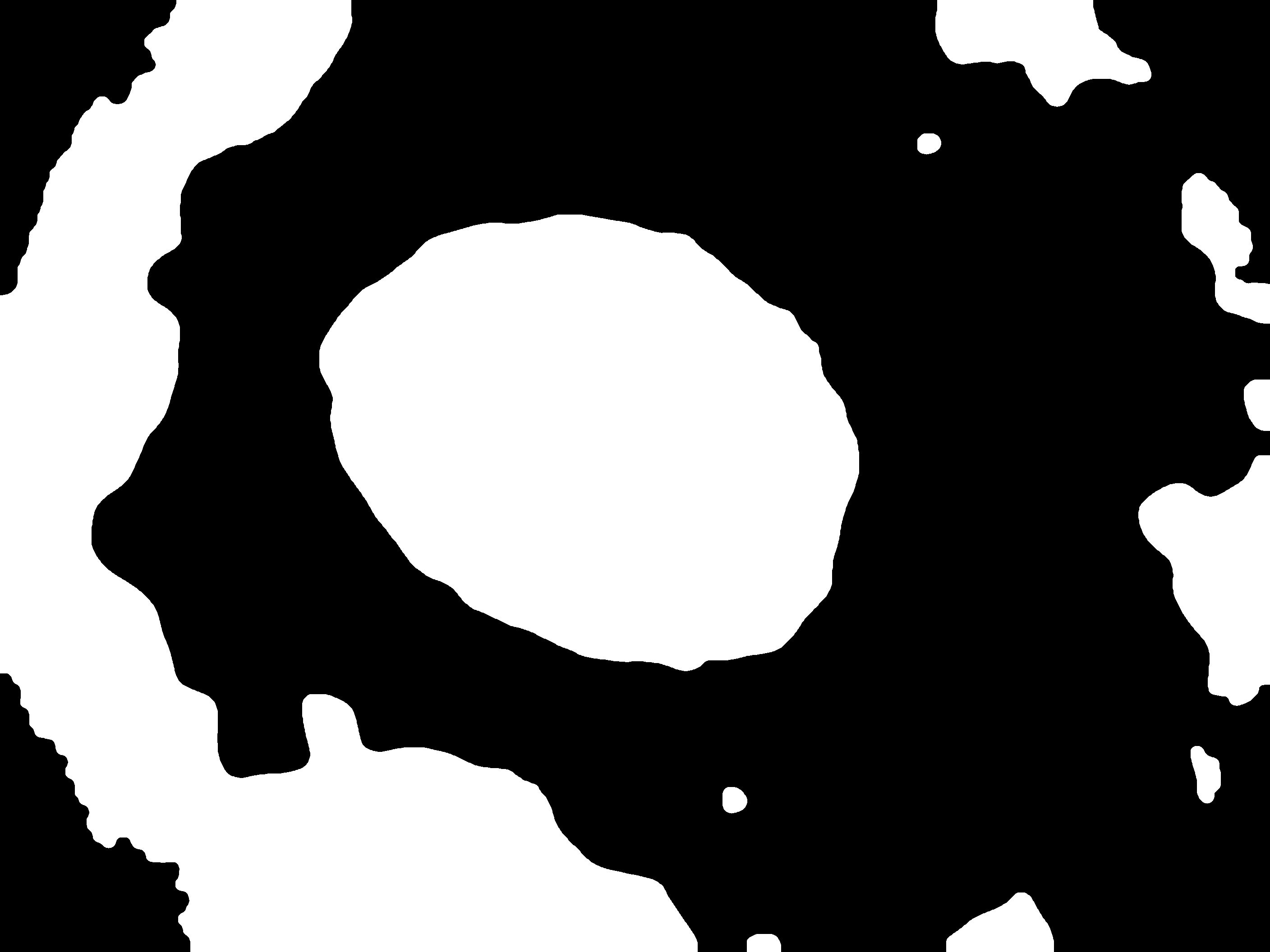}\\ 38.3\%}&
\multicolumn{1}{|p{1.8cm}|}{\centering 
\includegraphics[scale=0.018,valign=c]{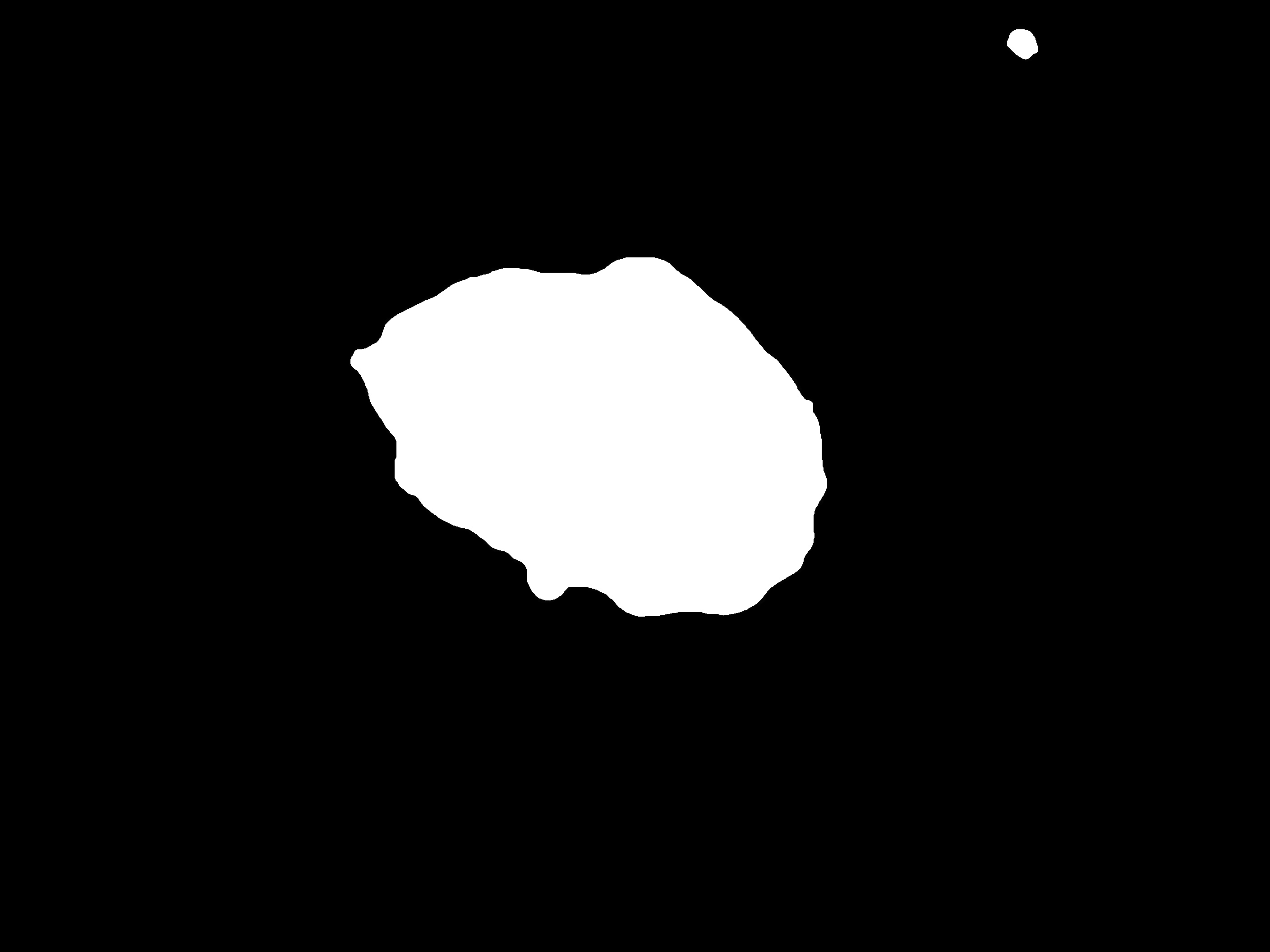}\\ 56.88\%} \\ \hline  

\includegraphics[scale=0.07,valign=c]{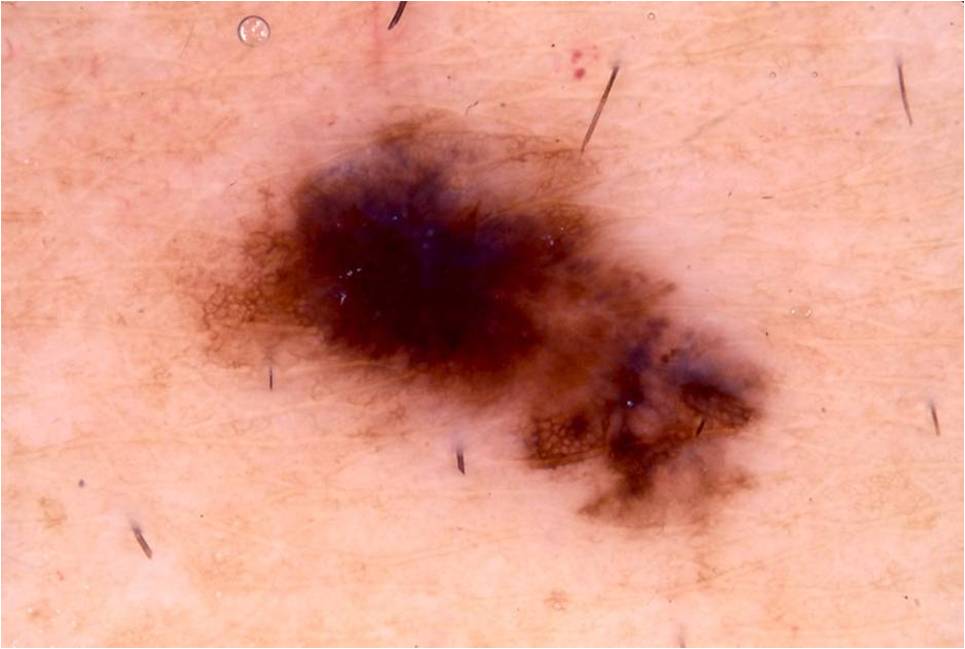}   & 
\includegraphics[scale=0.07,valign=c]{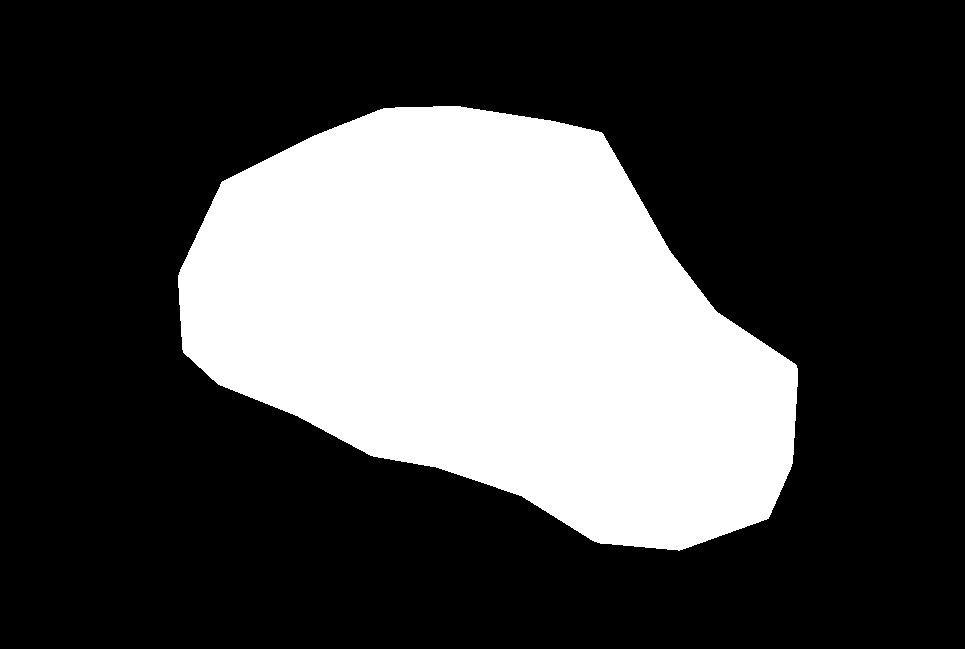}&  
\multicolumn{1}{|p{1.8cm}|}{\centering 
\includegraphics[scale=0.055,valign=c]{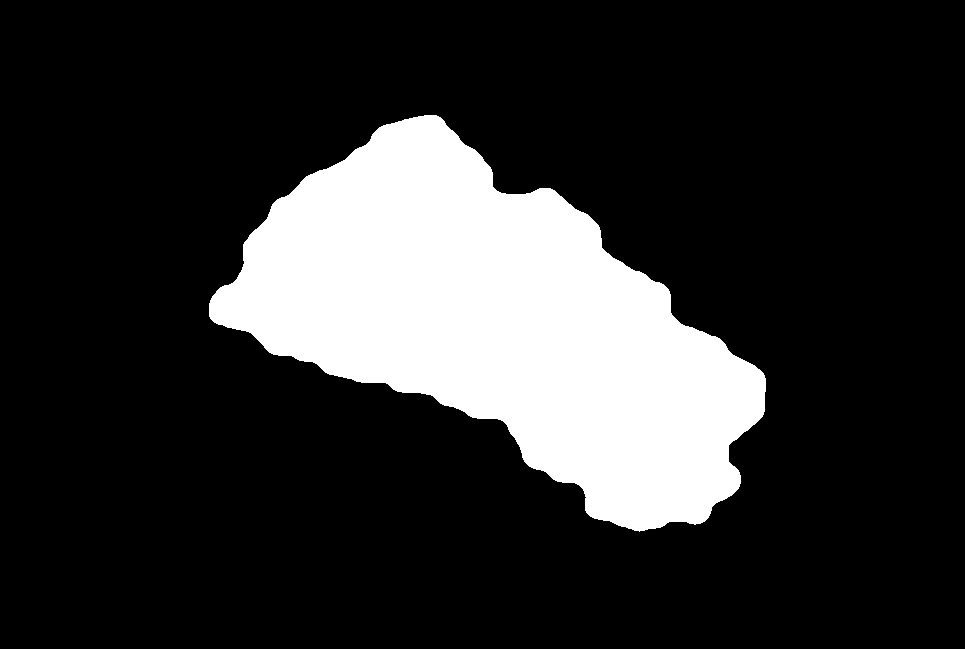}\\ 64\%}&
\multicolumn{1}{|p{1.8cm}|}{\centering 
\includegraphics[scale=0.055,valign=c]{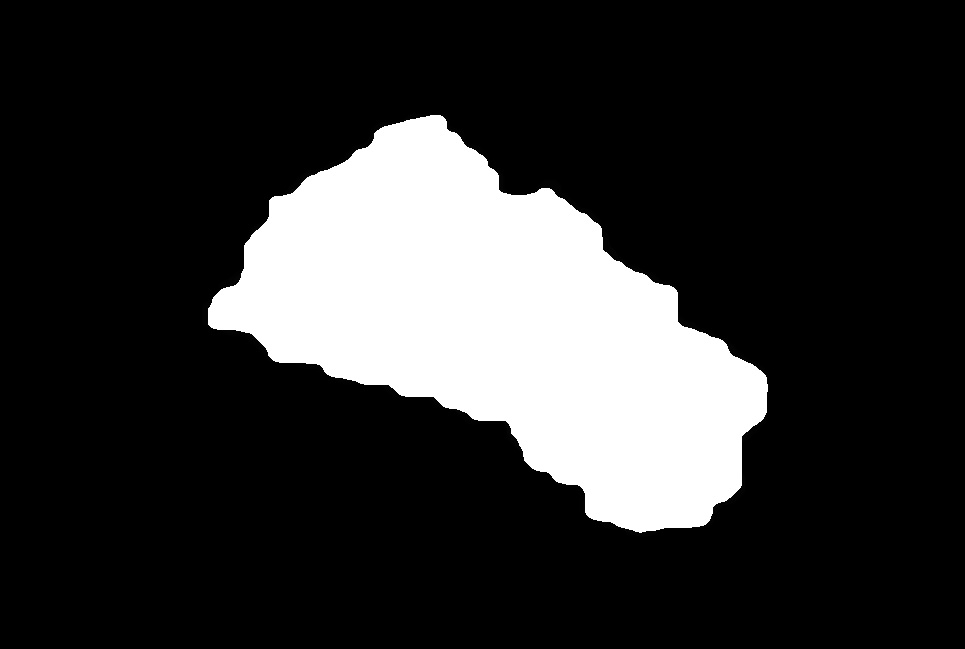}\\ 68\%} 
\end{tabular}
\caption{Comparing Gray-scale and BGR processing with method 2.}
\label{tab:method2_compare}
\end{table}
\subsection{Comparing both methods and Conclusion}
After extensive experiments, Method 1 showed noticeably higher Jaccards than Method 2. Table \ref{tab:comparing_results} shows 4 different examples.
The first row is an ordinary one with the lesion clearly recognized by both methods. The Second row shows a hairy image. Here, Method 2 outperforms method 1 due to the perfect hair removal techniques used. Row 3 is to show a microscopic example, that is, to see the effect of using Border filling, see Section \ref{sec:method1_pre-processing}. Lastly, row 4 shows a lesion with varying intensities. In the end, both methods were able to distinguish the vast majority of the 200 ISIC-2017 lesions with pretty good performances.

\label{sec:comparison_results}

\begin{table}[h]
\setlength\tabcolsep{1.5pt}
\begin{tabular}{c|c|c|c|}
Original &  Ground truth & method 1 & method 2  \\ \hline
\includegraphics[scale=0.14,valign=c]{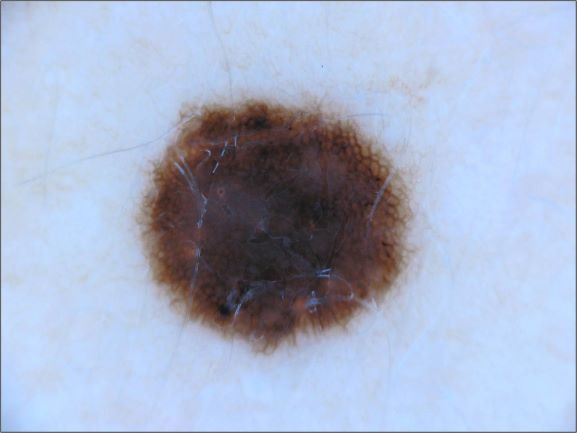}   & 
\includegraphics[scale=0.04,valign=c]{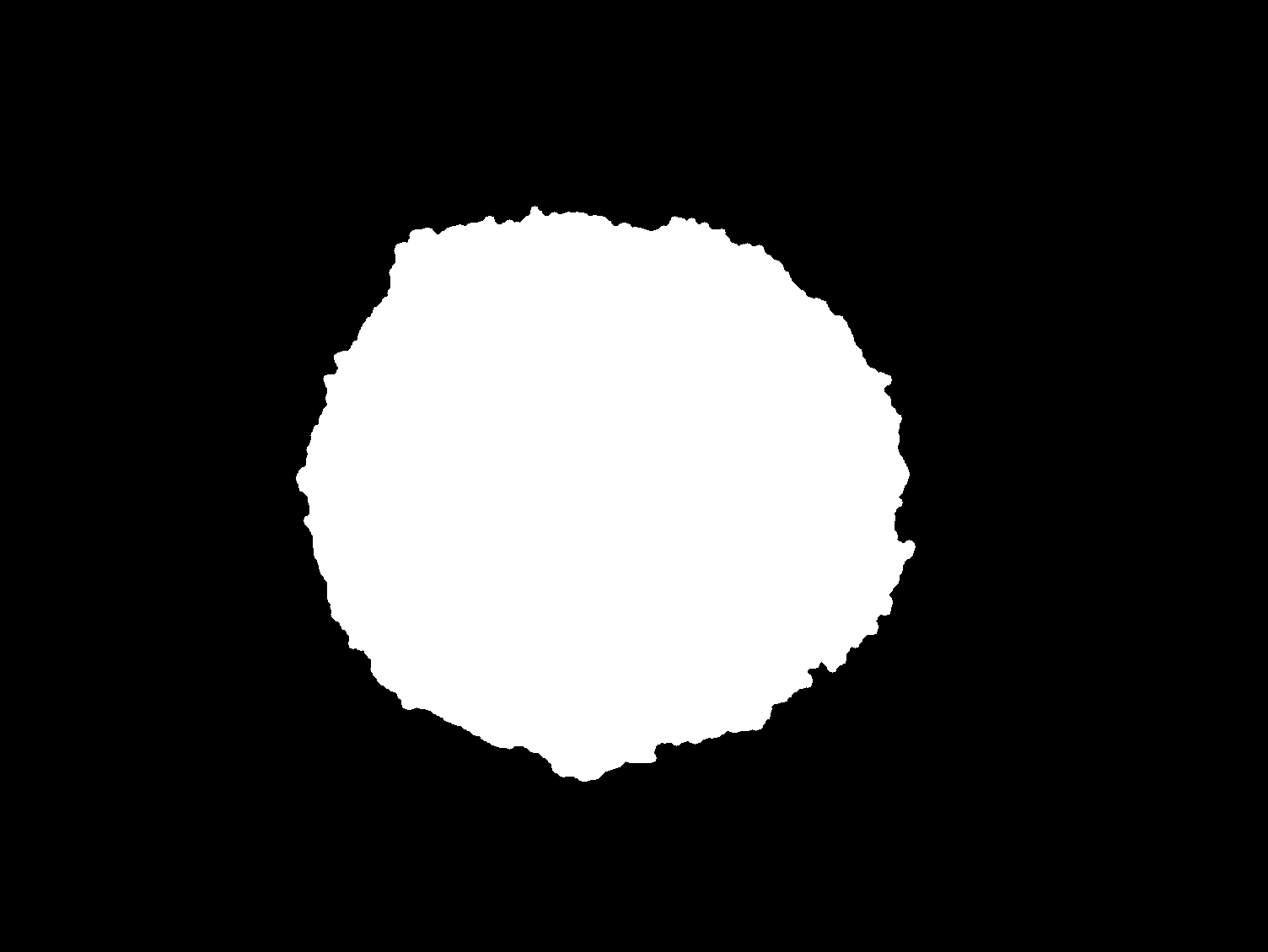}&  
\includegraphics[scale=0.04,valign=c]{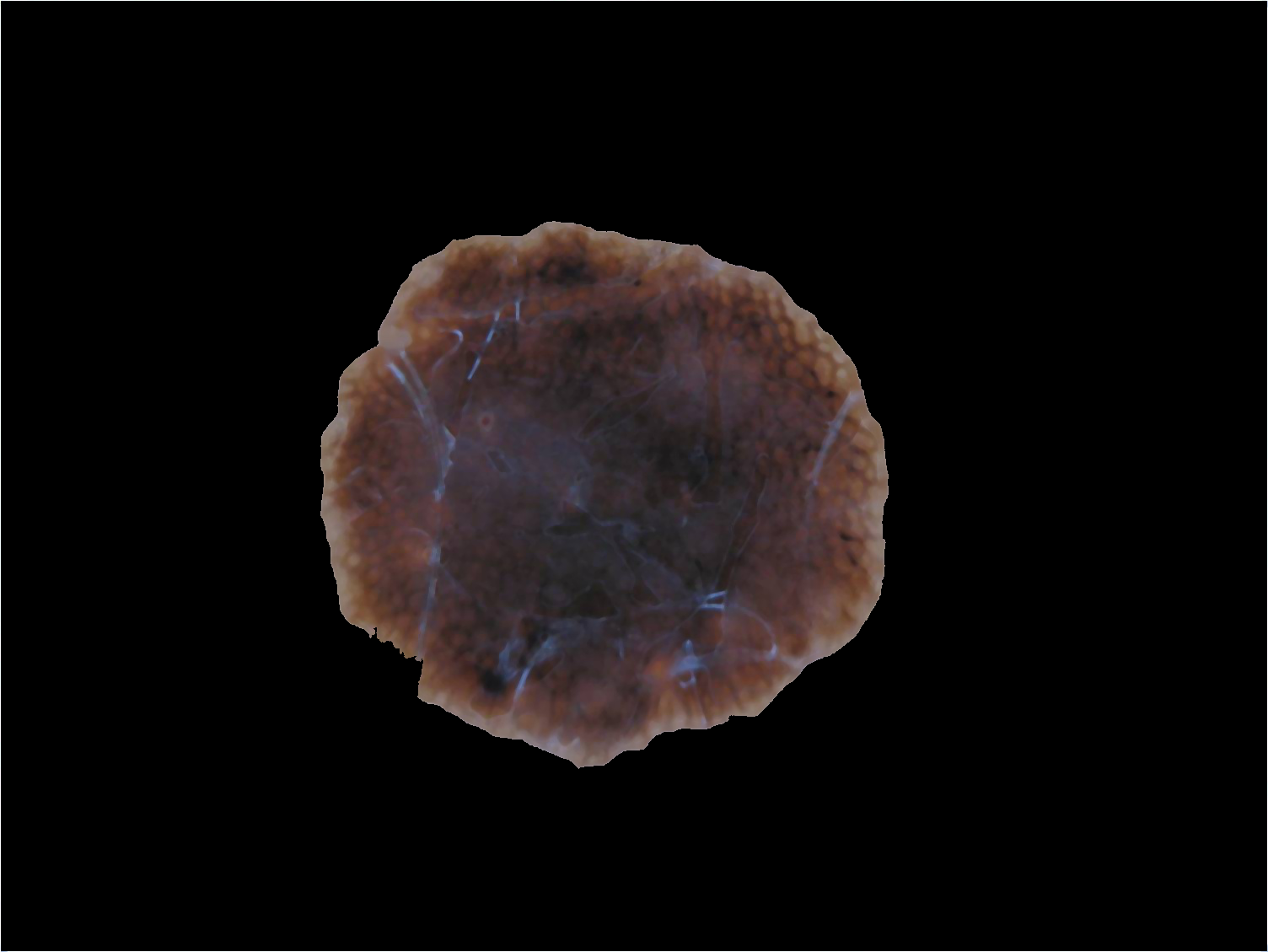}   &
\includegraphics[scale=0.04,valign=c]{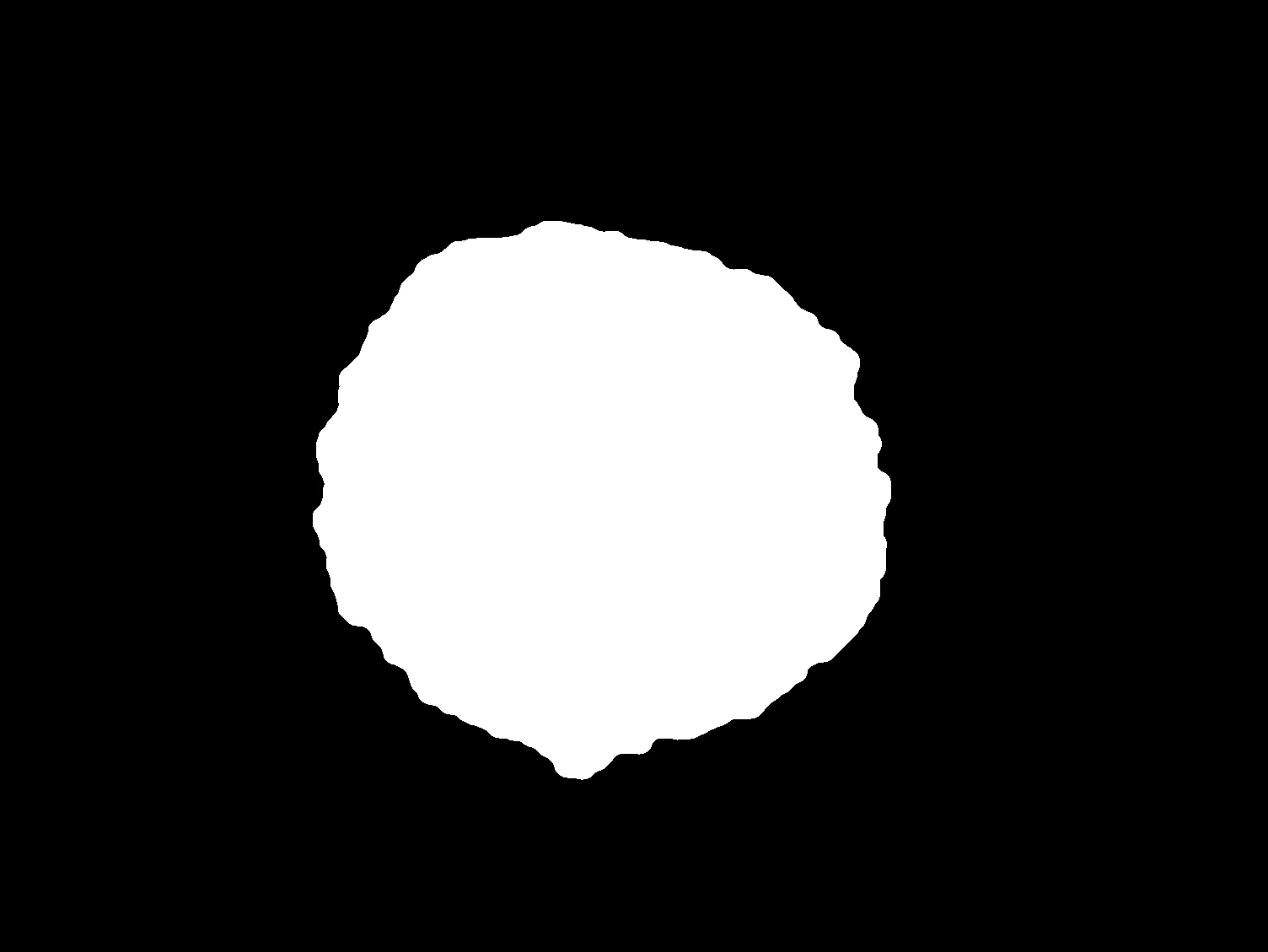}   
\\  \hline

\includegraphics[scale=0.14,valign=c]{images/ISIC_0000043.jpg}   &
\includegraphics[scale=0.04,valign=c]{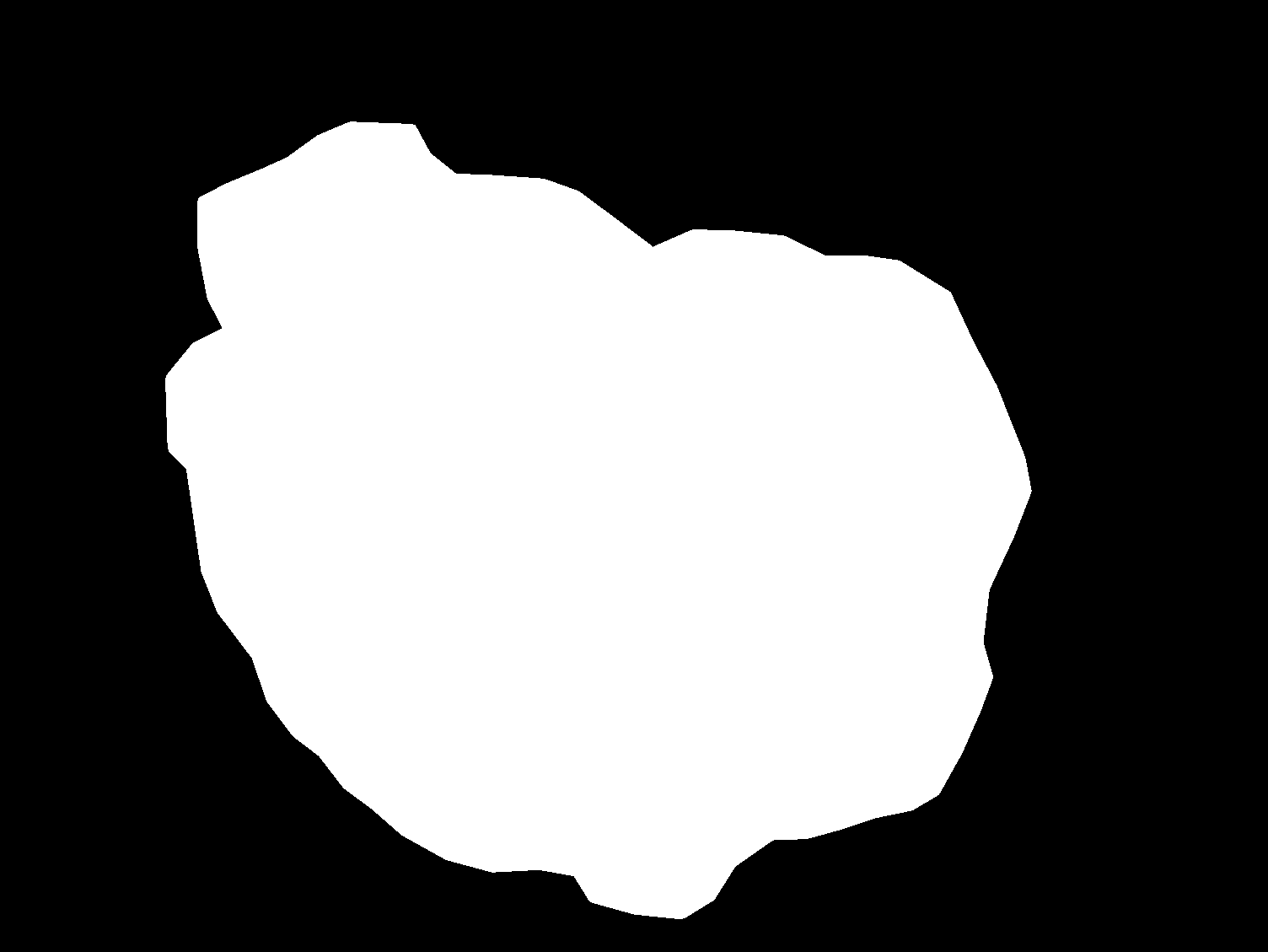}&  
\includegraphics[scale=0.04,valign=c]{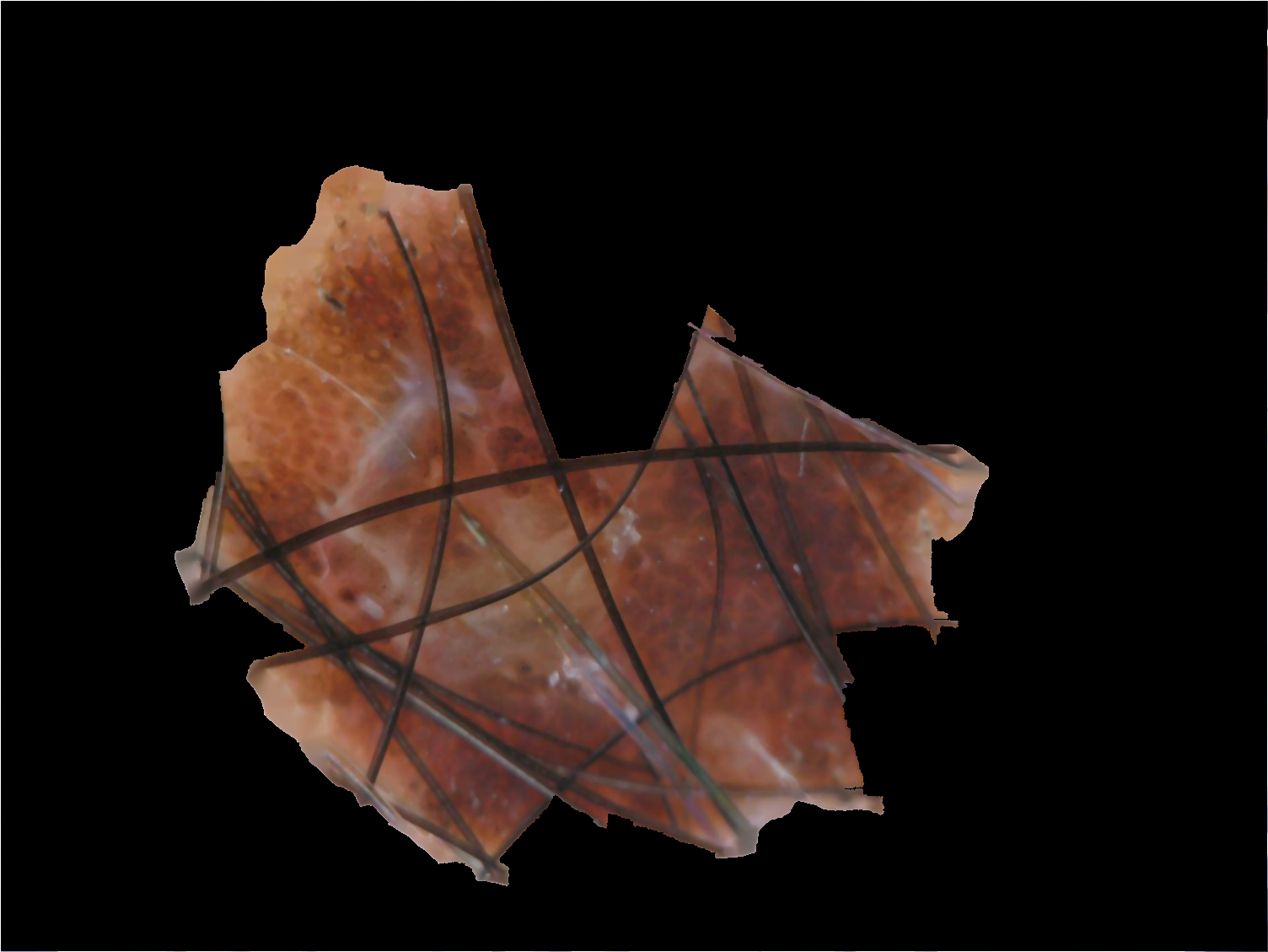}   &
\includegraphics[scale=0.04,valign=c]{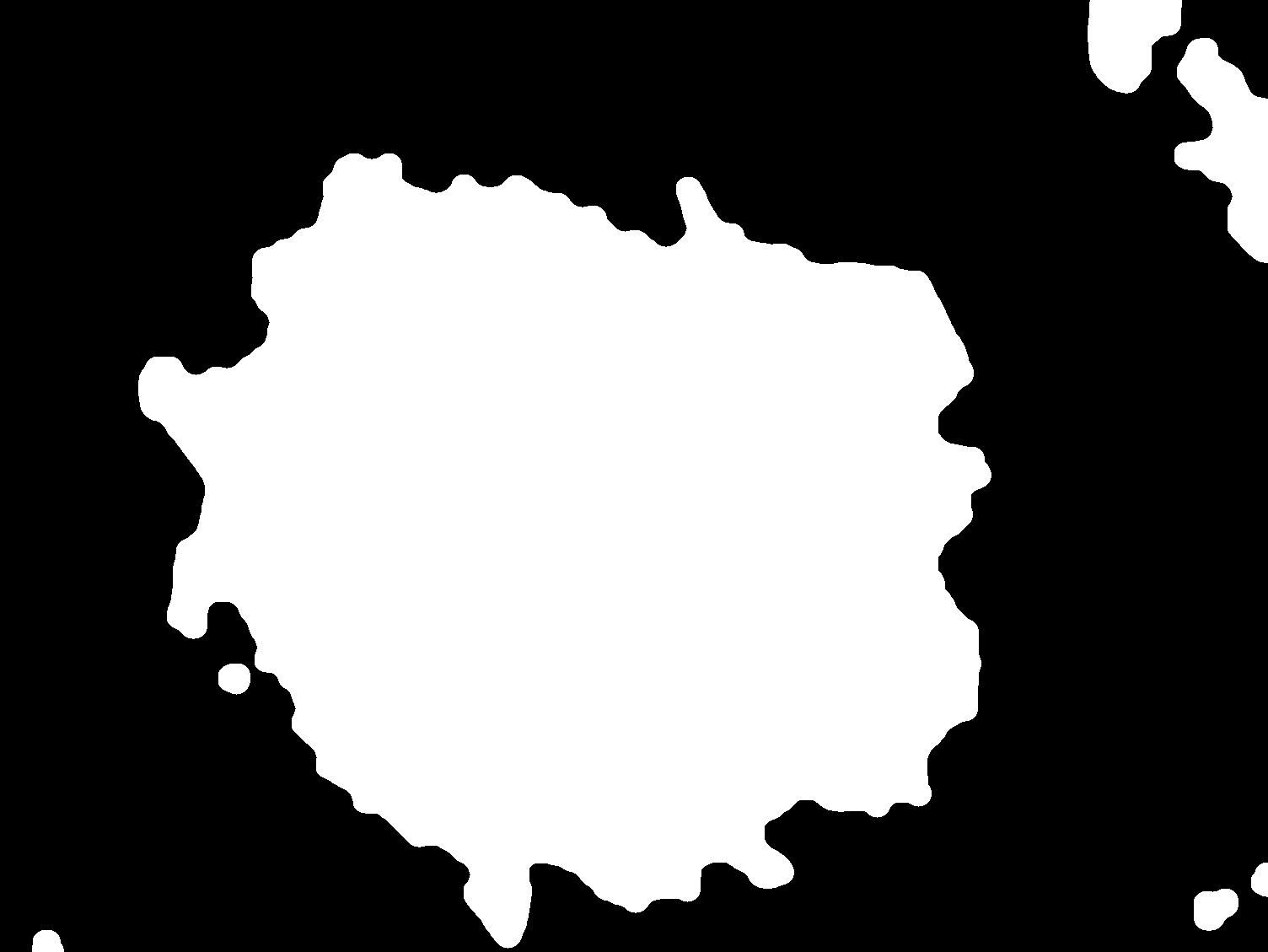}   
\\  \hline

\includegraphics[scale=0.14,valign=c]{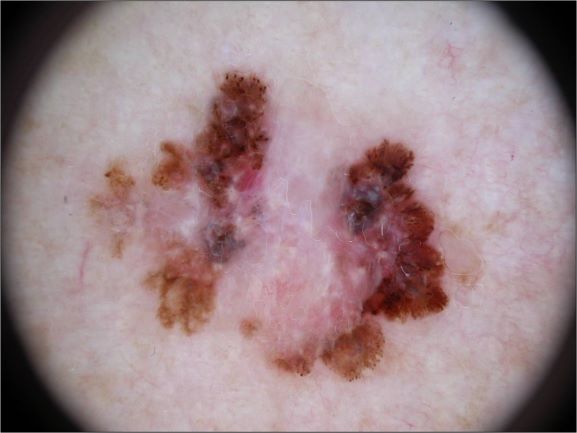}   & 
\includegraphics[scale=0.04,valign=c]{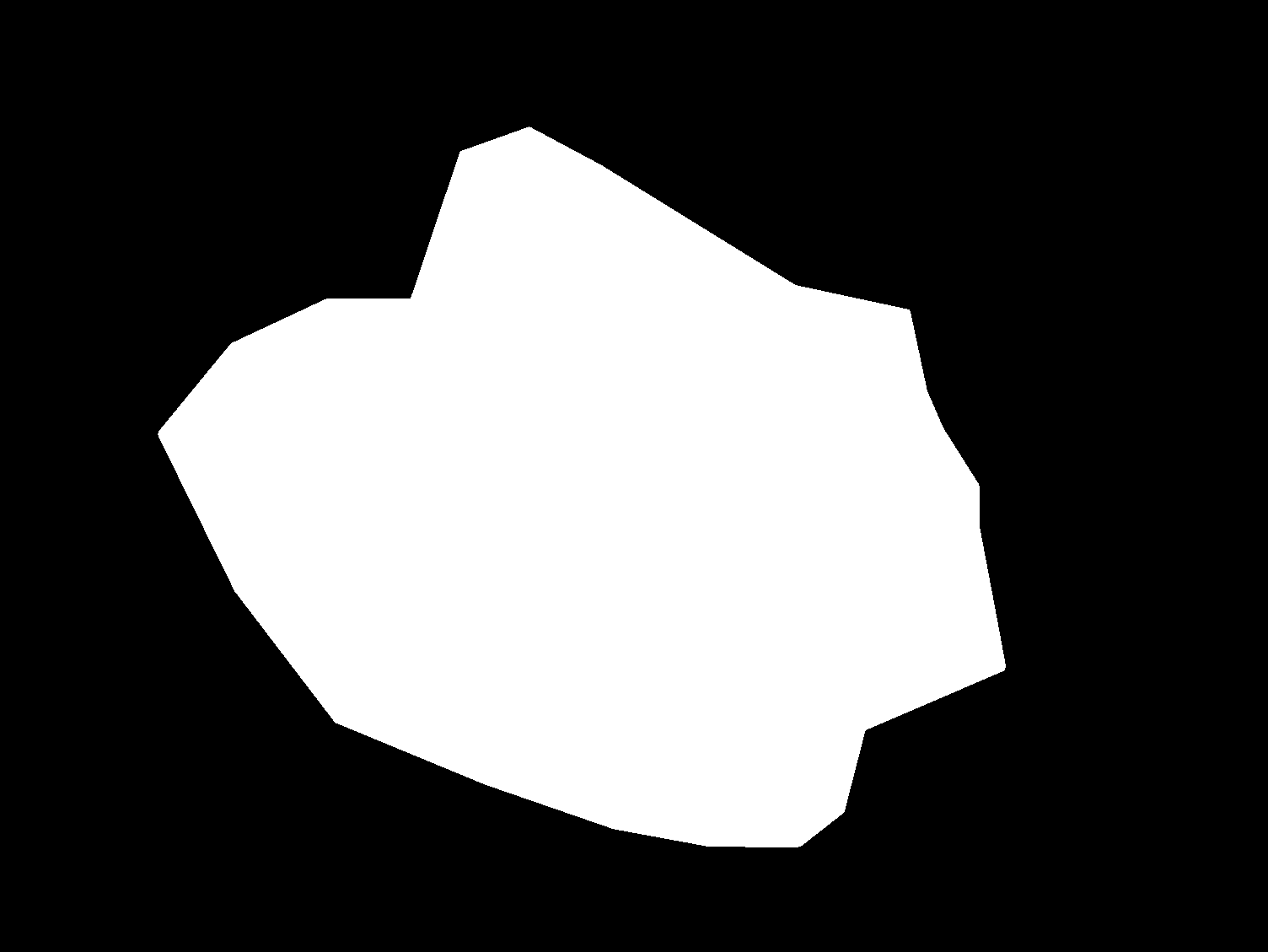}&  
\includegraphics[scale=0.04,valign=c]{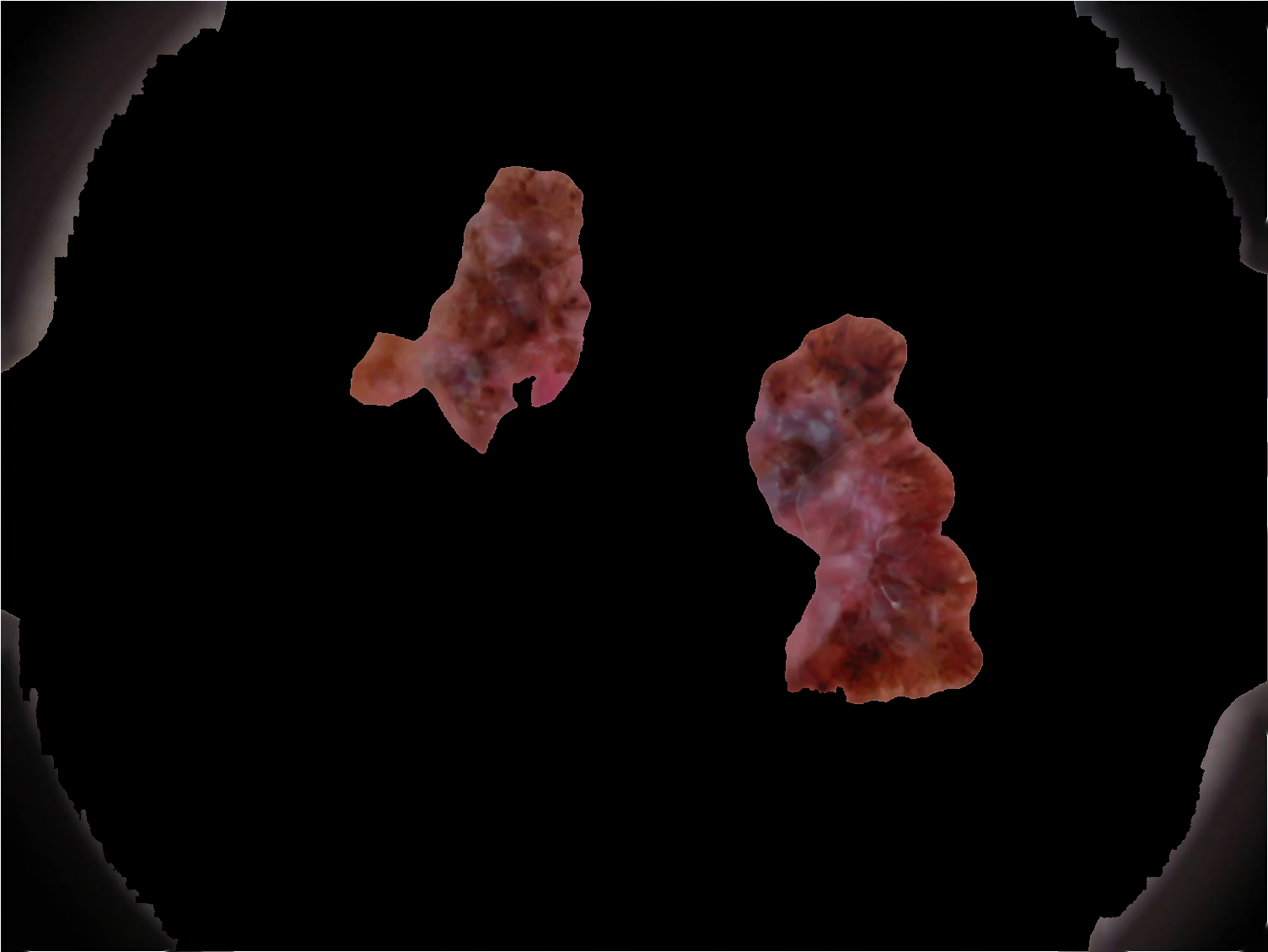}   &
\includegraphics[scale=0.04,valign=c]{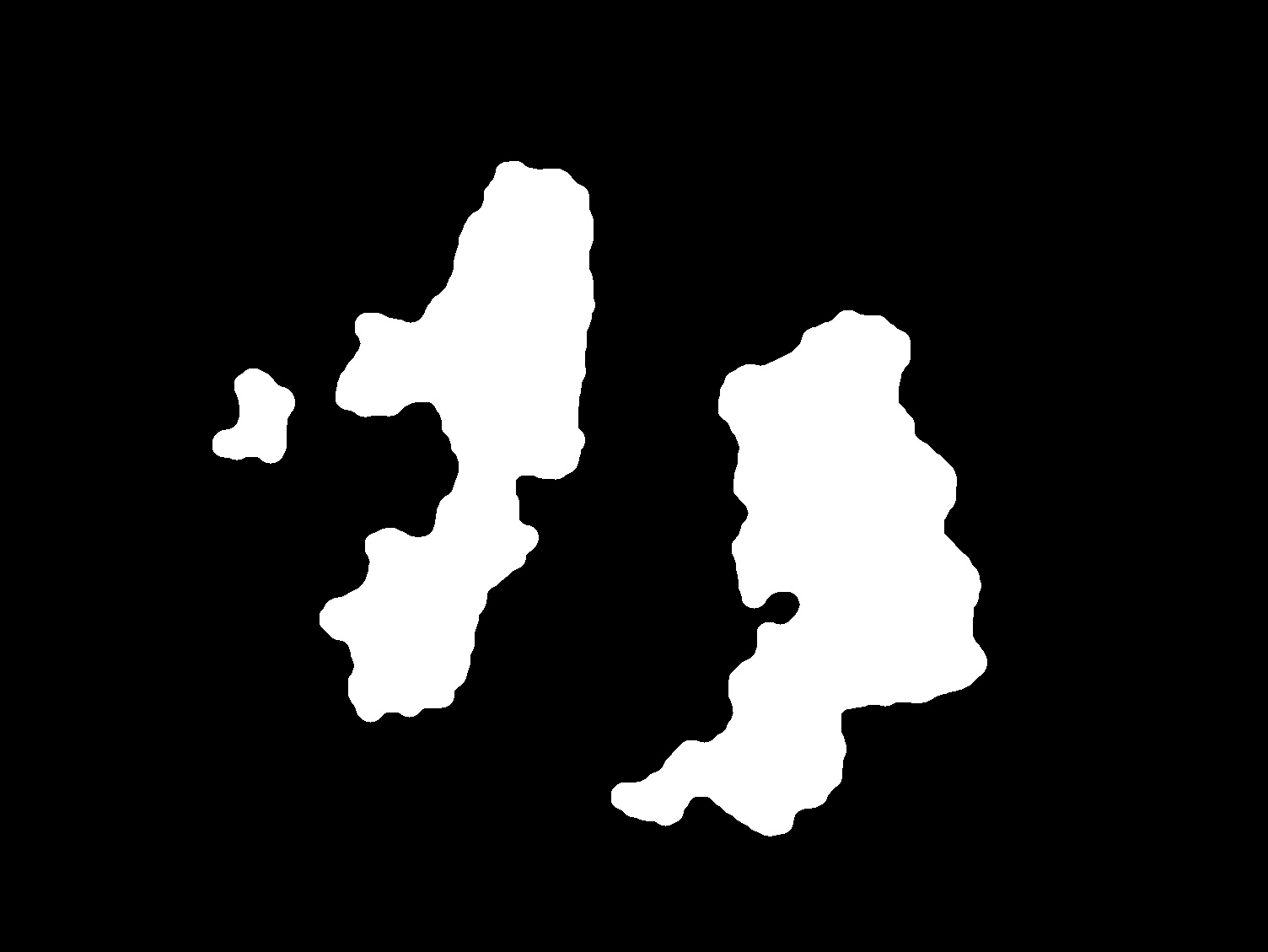}   
\\  \hline

\includegraphics[scale=0.14,valign=c]{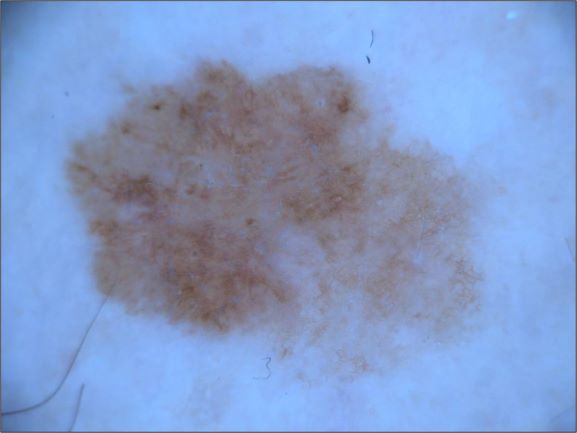}   & 
\includegraphics[scale=0.04,valign=c]{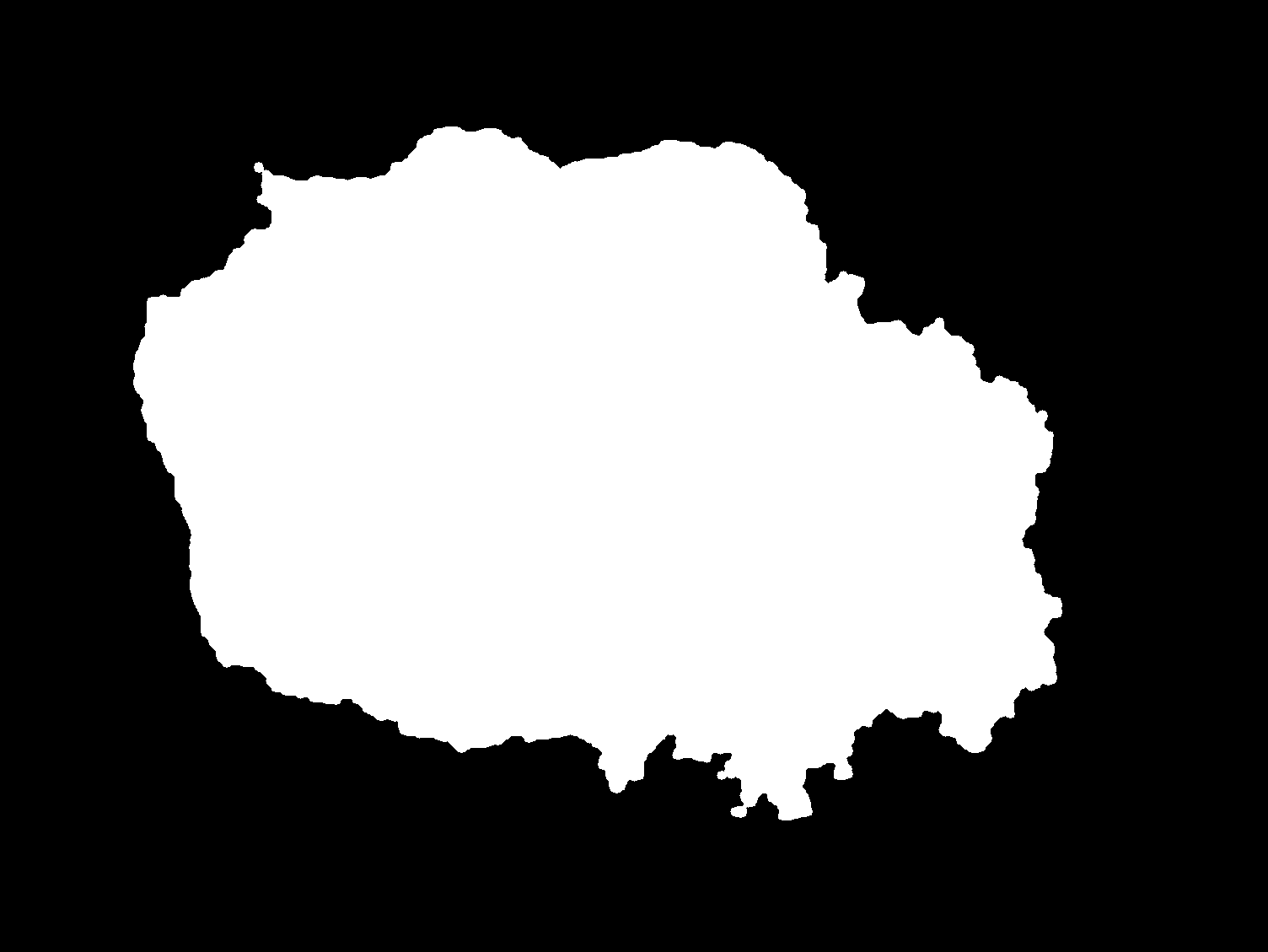}&  
\includegraphics[scale=0.04,valign=c]{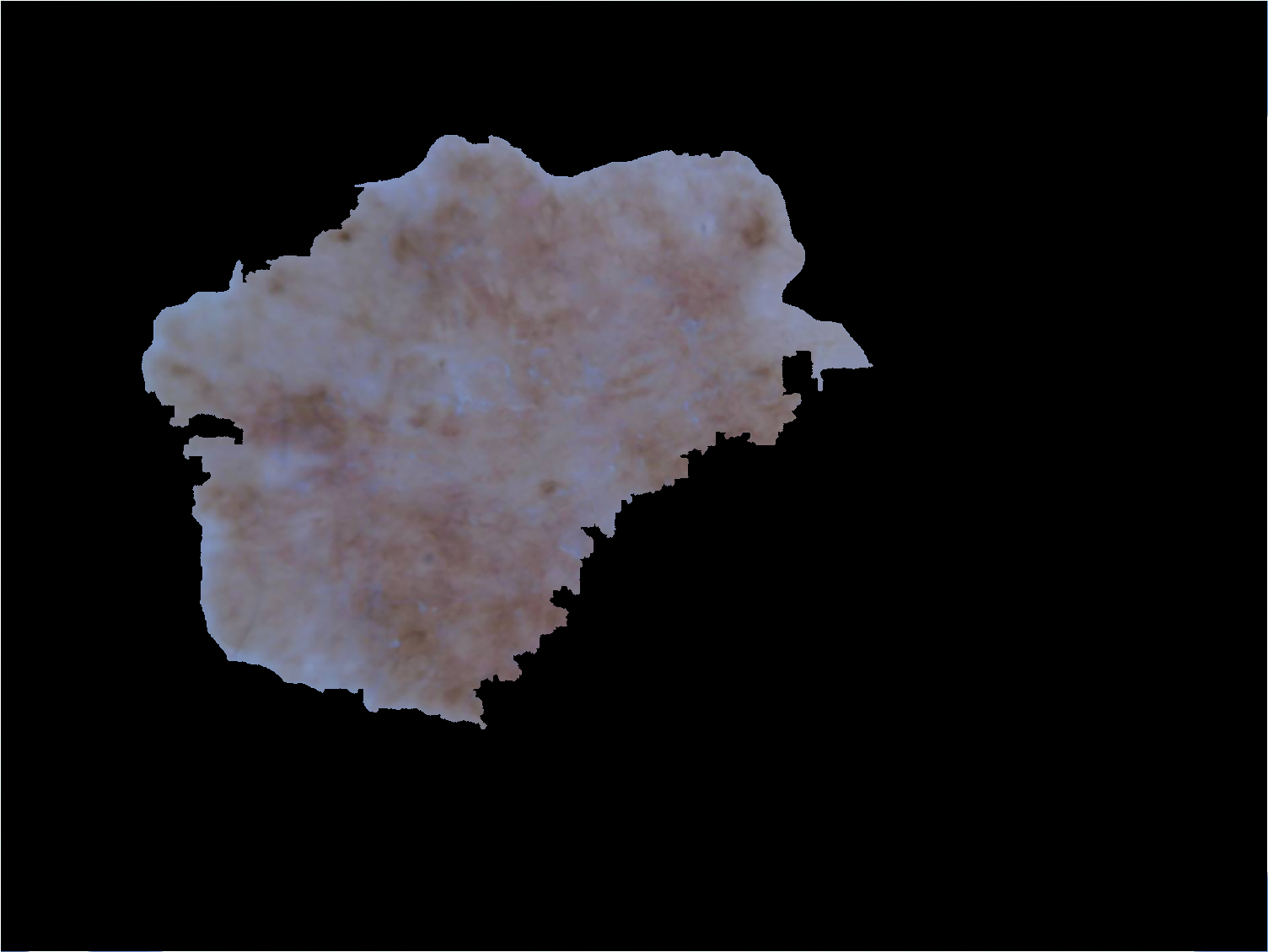}   &
\includegraphics[scale=0.04,valign=c]{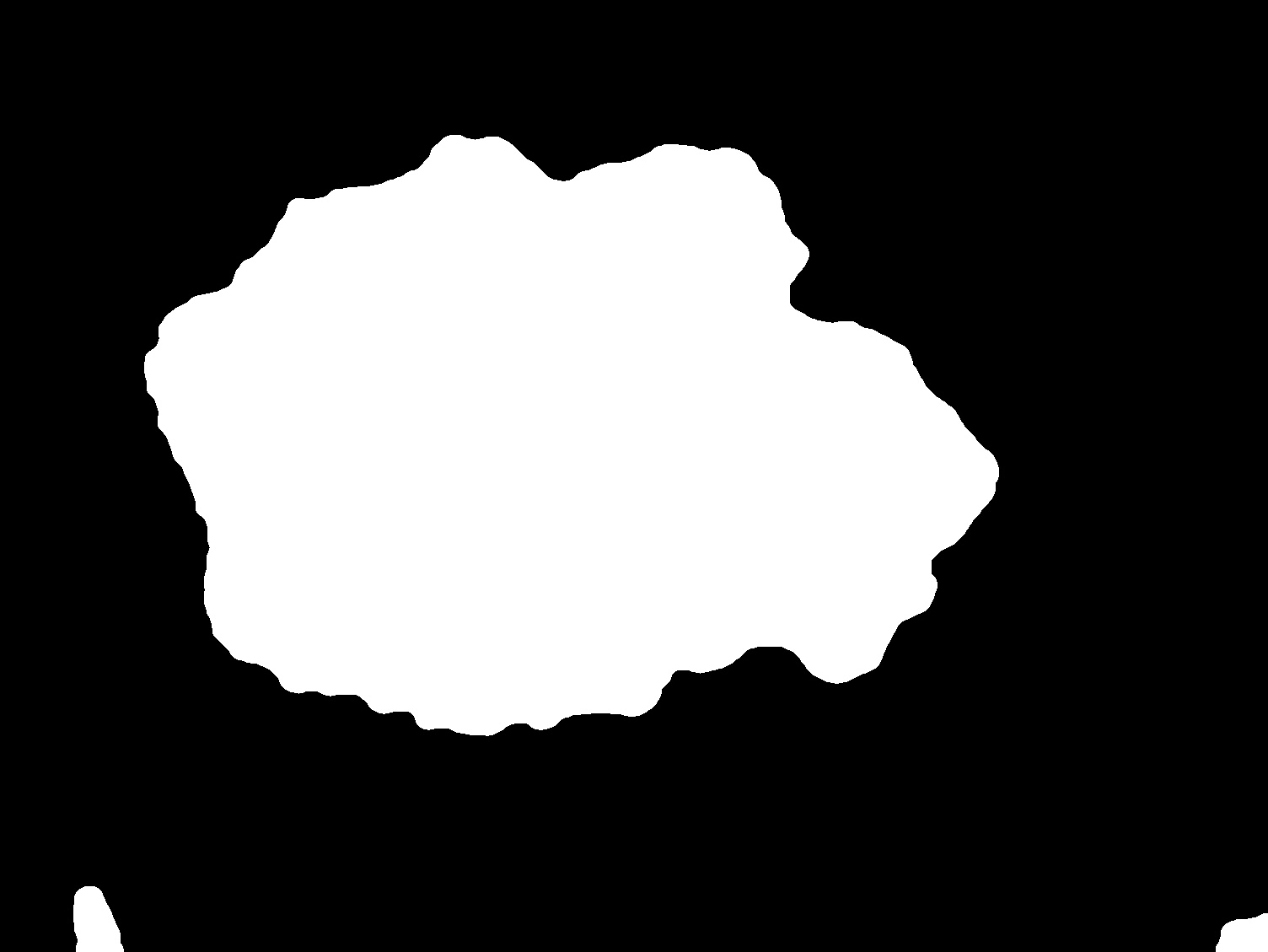}   
 \\  \hline

\end{tabular}
\caption{Results by both methods}
\label{tab:comparing_results}
\end{table}


\bibliographystyle{IEEEbib}


\end{document}